\pdfoutput=1

\documentclass[11pt]{article}

\usepackage{acl}

\usepackage{times}
\usepackage{latexsym}

\usepackage[T1]{fontenc}

\usepackage[utf8]{inputenc}

\usepackage{microtype}

\usepackage{inconsolata}

\usepackage{graphicx}

\usepackage{booktabs}
\usepackage{amsfonts}       
\usepackage{nicefrac}       
\usepackage{microtype}      
\usepackage{xcolor}         

\usepackage{makecell}
\usepackage{multirow}
\usepackage{adjustbox}
\usepackage{arydshln}
\usepackage[breakable]{tcolorbox}

\usepackage{listings}
\usepackage{enumitem}
\setlist{leftmargin=5.5mm}
\usepackage{subcaption}

\usepackage{amssymb}
\usepackage{pifont}

\definecolor{my_green}{RGB}{51,102,0}
\definecolor{my_yellow}{RGB}{255,165,0}
\definecolor{my_red}{RGB}{204, 0, 0}

\newcommand{\cmark}{\textcolor{my_green}{\ding{51}}} 
\newcommand{\gmark}{\textcolor{my_yellow}{\ding{51}}} 
\newcommand{\xmark}{\textcolor{my_red}{\ding{55}}} 
\definecolor{paperblue}{HTML}{077dea}
\definecolor{babyblue}{HTML}{E3EDF7} 

\newcommand{\coloredalpha}{\textcolor{paperblue}{\alpha}}
\newcommand{\coloredgamma}{\textcolor{paperblue}{\gamma}}
\newcommand{\coloredbeta}{\textcolor{paperblue}{\beta}}
\newcommand{\coloredsigma}{\textcolor{paperblue}{\sigma}}
\newcommand{\coloreddelta}{\textcolor{paperblue}{\delta}}
\newcommand{\coloredmu}{\textcolor{paperblue}{\mu}}
\newcommand{\coloredlambda}{\textcolor{paperblue}{\lambda}}
\newcommand{\coloredtau}{\textcolor{paperblue}{\tau}}

\NewDocumentCommand\cherrytitle{}{
    \includegraphics[scale=0.04]{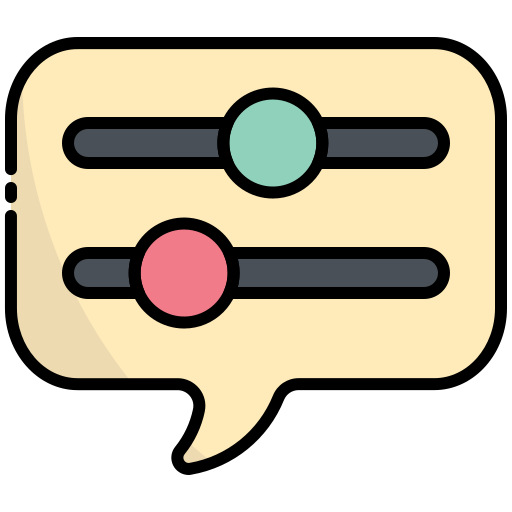}
}

\title{\cherrytitle Rethinking Stateful Tool Use in Multi-Turn \\ Dialogues: Benchmarks and Challenges}

\author{Hongru Wang$^{\coloredalpha}$, Wenyu Huang$^{\coloreddelta}$, Yufei Wang$^{\coloredgamma}$, Yuanhao Xi$^{\coloredsigma}$, Jianqiao Lu$^{\coloredmu}$, \\ \bf Huan Zhang$^{\coloredbeta}$, Nan Hu$^{\coloreddelta}$, Zeming Liu$^{\coloredlambda,}$, Jeff Z. Pan$^{\coloreddelta,}$\textsuperscript{\rm{\ding{41}}}, Kam-Fai Wong$^{\coloredalpha, \coloredtau,}$\textsuperscript{\rm{\ding{41}}} \\
  $^{\coloredalpha}$The Chinese University of Hong Kong,
  $^{\coloredgamma}$Macquire University,
  $^{\coloredlambda}$Beihang Univeristy \\
  $^{\coloreddelta}$The University of Edinburgh,
  $^{\coloredsigma}$Georg-August Universität Göttingen \\
  $^{\coloredmu}$The University of Hong Kong,
  $^{\coloredbeta}$Université de Montréal\&MILA \\
  $^{\coloredtau}$MoE Key Laboratory of High Confidence Software Technologies \\
  \texttt{\{hrwang, kfwong\}@se.cuhk.edu.hk}, \texttt{j.z.pan@ed.ac.uk} }

\begin{document}
\maketitle
\begin{abstract}
Existing benchmarks that assess Language Models (LMs) as Language Agents (LAs) for tool use primarily focus on stateless, single-turn interactions or partial evaluations, such as tool selection in a single turn, overlooking the inherent stateful nature of interactions in multi-turn applications. To fulfill this gap, we propose \texttt{DialogTool}, a multi-turn dialogue dataset with stateful tool interactions considering the whole life cycle of tool use, across six key tasks in three stages: 1) \textit{tool creation}; 2) \textit{tool utilization}: tool awareness, tool selection, tool execution; and 3) \textit{role-consistent response}: response generation and role play. Furthermore, we build \texttt{VirtualMobile} -- an embodied virtual mobile evaluation environment to simulate API calls and assess the robustness of the created APIs\footnote{We will use tools and APIs alternatively, there are no significant differences between them in this paper.}. Taking advantage of these artifacts, we conduct comprehensive evaluation on 13 distinct open- and closed-source LLMs and provide detailed analysis at each stage, revealing that the existing state-of-the-art LLMs still cannot perform well to use tools over long horizons.
\end{abstract}

\section{Introduction}
\label{sec:intro}

Large Language Models (LLMs) often rely on various tools to engage with external environments \citep{lu2023chameleon, zhuang2023toolqa}, in order to overcome their inherent limitations such as providing up-to-date information \citep{nakano2022webgpt} or domain-specific information \citep{li-etal-2023-api}, named tool learning \citep{qin2023tool, tool_tut}. Therefore, there are many previous studies that have been devoted to constructing benchmarks to evaluate the ability of LLMs to use different tools on various downstream environments/tasks \citep{zhuang2023toolqa, patil2023gorilla, mialon2023gaia, wang-etal-2024-appbench}. However, these efforts predominantly focus on stateless single-turn interaction, while overlooking the \textit{stateful tool use} in the multi-turn interactions \citep{zhuang2023toolqa, patil2023gorilla, mialon2023gaia, huang2024planning}. For instance, when a user fails to provide all the required arguments to use a tool in a single turn or requests details about a previous tool call, it becomes infeaible to provide detailed response without the tracking of tool states. In addition, most of existing benchmarks or environments fail to address the complexities of real-world interactions across the \textit{entire lifecycle of tool use}, encompassing tool creation, selection, execution, and integration of final responses, especially for tools with varying numbers and types of arguments \citep{li-etal-2023-api, qin2023toolllm}.

\begin{figure}
    \centering
    \includegraphics[trim={3.5cm 6cm 18cm 2cm}, clip, scale=1.0, width=0.49\textwidth]{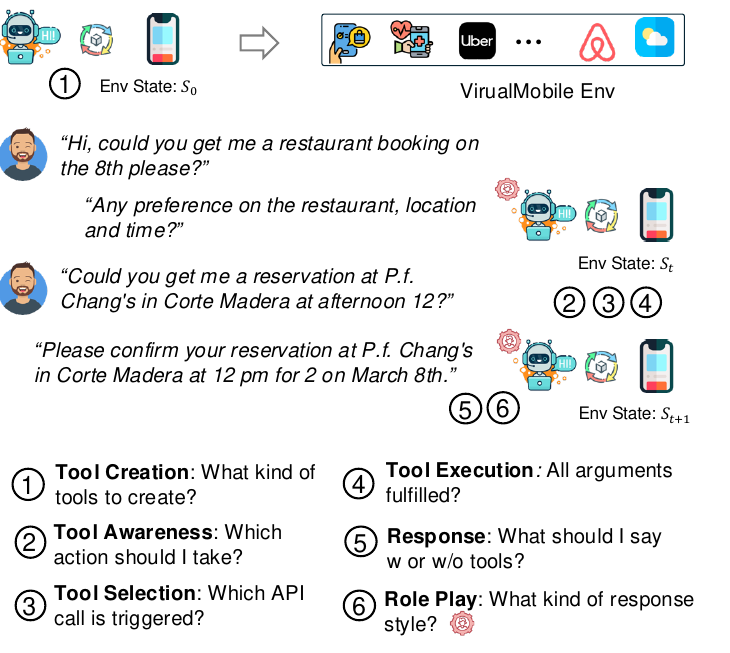}
    \caption{A typical example to show the entire life cycle of stateful tool use in multi-turn dialogues. The dialogue agent need to create the tools first or on the fly \ding{172}, and then decide whether or not use tools \ding{173}, which tool to use \ding{174}, execute it with all required arguments fulfilled \ding{175}, convert the tool results into responses with different role configs as conversion goes \ding{176}\ding{177}.}
    \label{fig:intro}
\end{figure}

\begin{table*}[!t]
\setlength{\belowcaptionskip}{0pt}
    \centering
    \begin{adjustbox}{max width=0.9 \textwidth}
    \begin{tabular}{l cccccc|cccc}
    \toprule
    \multirow{2}{*}{\textbf{Benchmark}} & \multicolumn{6}{c}{\textbf{Tool Learning}}  & \multicolumn{4}{c}{\textbf{Evaluation}} \\

    \cline{2-11} & Apps & APIs & Argu. & C. S. E. & States & Awareness & Role &  Hierarchical & Resp. & Multi-turn \\
    \hline
    APIBench \citep{patil2023gorilla} & 3 & 1,715 & (1.5/1.0) & \xmark | \gmark | \cmark & \xmark & \xmark & \xmark & \xmark & \xmark & \xmark \\
    API-Bank \citep{li-etal-2023-api} & 8 & 53 & (2.5/1.0) & \xmark | \cmark | \cmark & \xmark & \xmark & \xmark & \xmark & \cmark & \cmark \\
    ToolBench \citep{qin2023toolllm} & 49 & 16,464 & (1.0/1.0) & \xmark | \gmark | \cmark & \xmark & \xmark & \xmark & \cmark & \xmark & \cmark \\
    ToolQA \citep{zhuang2023toolqa} & 6 & 13 & (1.0/1.0) & \xmark | \gmark | \cmark & \xmark & \xmark & \xmark &  \xmark & \cmark & \xmark \\
    GAIA \citep{mialon2023gaia} & - & - & - & \xmark | \cmark | \cmark & \xmark & \xmark & \xmark & \xmark & \cmark & \xmark \\
    UltraTool \citep{huang2024planning} & 22 & 2032 & (4.1/1.6) & \cmark | \cmark | \cmark & \xmark & \xmark & \xmark & \xmark & \xmark & \xmark \\
    
    \hline
    
    AgentBench \citep{liu2023agentbench} & 8 & -  & - & \xmark | \gmark | \cmark & \xmark & \xmark & \xmark &  \xmark & \xmark & \xmark \\
    MINT \citep{wang2024mint} & 8 & - & - & \xmark | \cmark | \cmark & \xmark & \xmark & \xmark & \xmark & \cmark & \xmark \\
    AgentBoard \citep{ma2024agentboard} & 9 & - & - & \xmark | \cmark | \cmark & \cmark & \xmark & \xmark & \xmark & \xmark  & \cmark \\
    \hline
    \hline
    DialogTool & 16 & 31$^{\heartsuit}$ & (\textbf{4.2}/\textbf{7.5}) & \cmark | \cmark | \cmark & \cmark & \cmark & \cmark & \cmark & \cmark & \cmark \\
    \bottomrule
    \end{tabular}
    \end{adjustbox}
    \caption{Comparison with existing evaluation benchmarks (first part: tool learning benchmarks; second part: agent benchmarks) where the C.S.E. stands for Creation | Selection | Execution of tools, and \gmark stands for the selection of tool does not consider the case which does not need any tools. $^{\heartsuit}$ We emphasize that we focus on the interaction and complexity of each API instead of solely number of APIs. Thus we list the average number of input and returned arguments by APIs in Argu. Culumn. Hierarchical stands for hierarchical tool design in our \texttt{VirtualMobile} in terms of App, API and Arguments.}
    \label{tab:existing_work}
\end{table*}

To maintain seamless interaction over long horizons, we introduce \texttt{DialogTool}, the first benchmark designed to comprehensively evaluate the entire lifecycle of stateful tool use in multi-turn dialogues. Table~\ref{tab:existing_work} compares \texttt{DialogTool} with existing benchmarks. Generally, we leverage existing dialogue datasets, particularly task-oriented dialogue datasets (TDD) \citep{budzianowski-etal-2018-multiwoz, rastogi2020towards}, to gather data and construct the corresponding evaluation environment efficiently and effectively. In detail, on the data side, we regard the \textit{service/domain}, \textit{slots} and \textit{intents} in TDD as different \textit{Apps}, \textit{Arguments}, and \textit{APIs}, and transform every database lookup operation in the dialogue into an API function call, adhering to the standard tool call paradigm \citep{li-etal-2023-api, wang-etal-2024-appbench}. On the environment side, we firstly store the output for each API call as the database, and then manually implement each function for all APIs and ensure the correctness\footnote{Given same input in the dialogue, it can produce same output, }, resulting in a virtual mobile environment (\texttt{VirtualMobile}) with lots of supported Apps and APIs. For example, the user may want to find one restaurant with specific food and location, and the result can be returned using \colorbox{lightgray!30}{FindRestaurant} API in \colorbox{lightgray!30}{Restaurant} App that takes the desired food type and location as input parameters, and returns a list of names of matching restaurant. 


Building on top of \texttt{DialogTool} and \texttt{VirtualMobile}, we can assess the entire lifecycle of stateful tool use by examining six dimensions across three different stages (Figure~\ref{fig:intro}): 1) \textbf{Tool Creation} to generate code function given the whole tool description; 2) \textbf{Tool Utilization} which consists of tool awareness to determine whether or not require tools, tool selection to select appropriate API and tool execution to fulfill all arguments; and 3) \textbf{Role-consistent Response} to generate final responses according to different roles (i.e., role play) and tool states (i.e., response generation). It is worth noting here that the role playing transforms responses into different styles to enhance user engagement, independent of the tools being used, allowing for varied expressions regardless of the specific tools employed. To conclude, our contributions can be summarized below:

\begin{itemize}[leftmargin=*,topsep=1pt,itemsep=1pt]
    \item To the best of our knowledge, this is the first attempt to evaluate the whole life cycle of stateful tool use in the context of multi-turn dialogue, including six key dimensions across three distinct stages.

    \item We propose \texttt{DialogTool}, the first multi-turn dialogue dataset considering stateful and interactive tool use, accompanying with an embodied virtual mobile environment (\texttt{VirtualMobile}), ensuring the reproducibility and evaluation of different LLMs to interact with both humans and APIs over long horizons.

    \item We conduct extensive experiments on 13 distinct LLMs of varying sizes, covering both state-of-the-art open- and closed-source models, and then provide comprehensive analysis in each stage of tool use and address the challenges encountered in practice.
\end{itemize}

\section{Related Work}

\paragraph{Task-oriented Dialogue.} Task-oriented Dialogue Systems (DS) have undergone significant transformations with the progress of Language Models (LMs) \citep{wang2023survey}. Despite the differences in models, the core of a dialogue system lies in determining the next action and coordinating various knowledge from different services to complete the task. For example, \citet{rastogi2020towards} propose a Schema-Guided Dialogue (SGD) dataset considering an ever-increasing number of services and APIs spanning multiple domains. However, the majority of current dialogue systems lean towards traditional frameworks while utilizing LLMs as their foundational models \citep{hudecek-dusek-2023-large, zhang-etal-2023-sgp}, benefiting from well-established techniques. In contrast, several advancements have begun to explore tool learning in dialogue system, treating different APIs \citep{shu2022dialog2api, li-etal-2023-api} or knowledge sources \citep{wang-etal-2023-large} as individual tools. Building on this, recent studies have increasingly focused on complex tool interactions under specific constraints, such as domain policies \citep{yao2024taubench}, drawing inspiration from traditional task-oriented dialogue research \cite{xu-etal-2024-rethinking, lu2024toolsandbox}. 





\noindent \textbf{Tool Learning.} Tools have been defined as cognitive tools or physical tools \citep{tool_tut}, where the former is defined as a cognitive concept inside human beings comes from cognitive science \citep{gigerenzer1991tools,baron1991precursors, wang-etal-2023-cue} and the later comes from external physical world such as different models \citep{shen2023hugginggpt}, search engine~\citep{wang2024selfdc}, APIs \citep{li-etal-2023-api, wang-etal-2024-appbench}, and even robot manipulation \citep{huang2022language, liang2023code}. Most of previous works, such as APIBench \citep{patil2023gorilla} and ToolQA \citep{zhuang2023toolqa}, have primarily revolved around the selection and execution of tools. This includes tasks such as identifying the right tool for a given instruction and understanding all the necessary arguments needed to execute the determined tool. Furthermore, \citet{mialon2023gaia} consider the cases which do not require tools and require multiple tools in single turn. AgentBoard \citep{ma2024agentboard} consider the progress evaluation at each step to complete the complex task. Distinguishing from these previous works, we focus on the whole life cycle of tool utilization across three distinct stages, and introduce more fascinating features such as role playing and practical hierarchical structure \footnote{More related work can be found in Appendix.}.

\noindent \textbf{Role Play.} Assign LLMs some pre-defined roles has been proved an effective way to engaging user \citep{wang2023rolellm}, resulting in more longer interaction time , such as character.ai\footnote{https://character.ai/}. Most existing work focuses on character roles rather than assistant roles. For example, CharacterEval \citep{tu2024charactereval} evaluates LLMs on generating role-consistent responses based on a given role background. In contrast, we focus on the language style of different roles \citep{zhou2023characterglm}, aiming to generate more user-friendly and preferable responses.

\section{Dataset and Environment}

\subsection{Seed Dataset}
To create our \texttt{DialogTool} dataset effectively and efficiently, we prioritize using existing task-oriented dialogue datasets (TDD) that closely resemble real-world interactions while minimizing human effort. We first select seed datasets based on two main criteria: 1) The datasets should well reflect how tools or functions are invoked as the conversation goes, such as dialogue system call different APIs in the multi-turn task-oriented dialogue dataset; 2) We prefer datasets that offer diverse conversations with comprehensive and detailed annotations, ranging from different domains and tools Specifically, we mainly incorporate the SGD dataset \citep{rastogi2020towards} and also MultiWoZ \cite{budzianowski-etal-2018-multiwoz,zang-etal-2020-multiwoz} due to their extensive coverage across various domains, slots, and slot values, featuring a dynamic ontology of APIs spanning numerous domains. 


\begin{figure}
    \centering
    \includegraphics[trim={2cm 2cm 8cm 1cm}, clip, scale=1.0, width=0.5\textwidth]{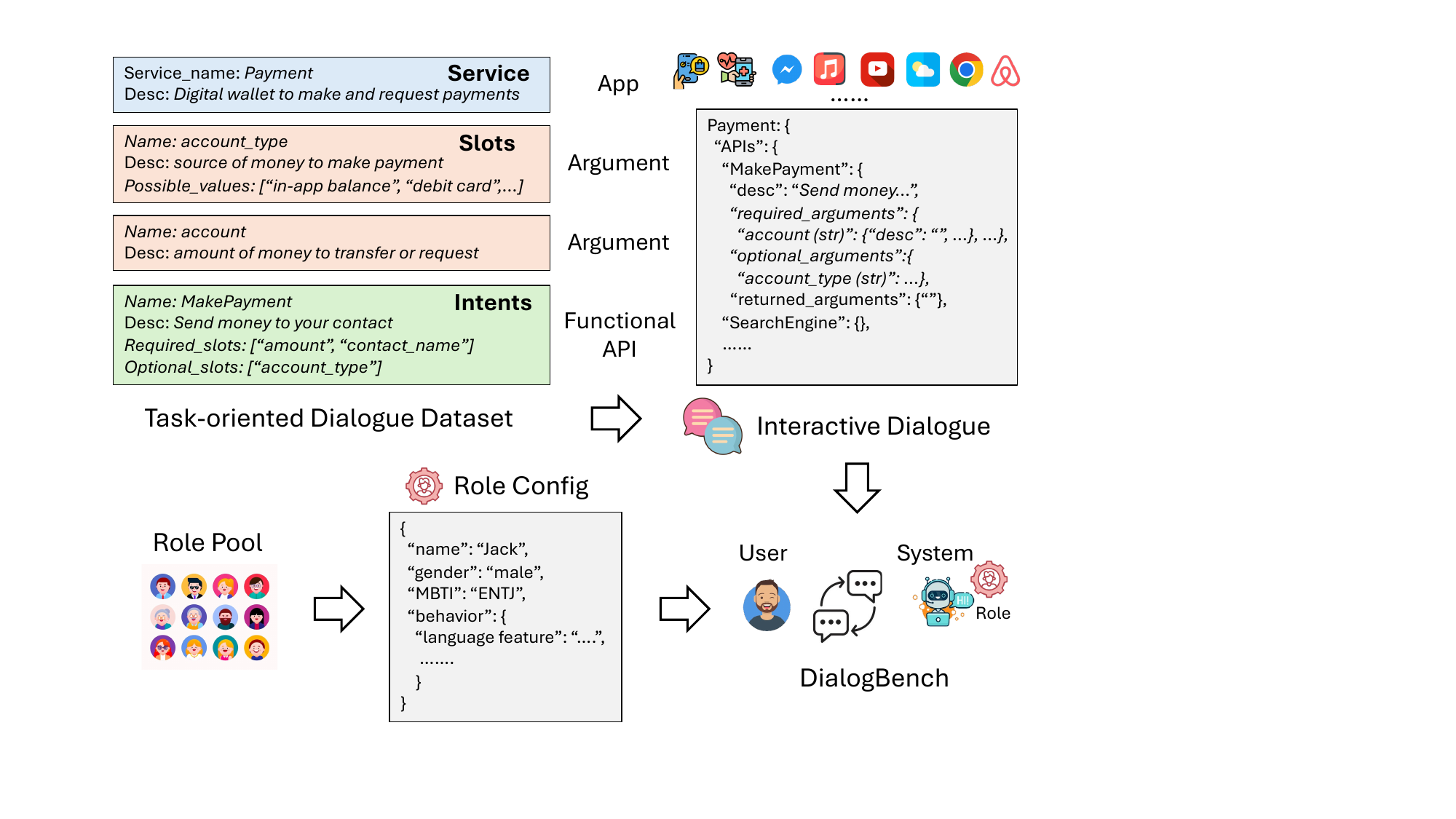}
    \caption{The pipeline of \texttt{DialogTool} collection by 1) \textbf{Setting shift}: transfer the setting of existing dialogue datasets; 2) \textbf{Role Play}: then rewriting the utterances from system side into role-consistent utterances.}
    \label{fig:data_collection}
\end{figure}

\subsection{Dataset Collection}
\label{sec:dataset_collection}

Figure~\ref{fig:data_collection} shows the details of data collection. In detail, there are two steps: 1) \textit{Setting Shift}: we write a Python script to automatically transform the annotations of task-oriented dialogue dataset (except the utterances), into conventional labels in tool learning (i.e., tool selection and execution); 2) \textit{Role Playing}: In order to provide role-consistent response, we assign dialogue agent different roles, resulting in varied response styles.


\paragraph{Setting Shift.} Lots of recent studies try to build tool learning benchmark from scratch \cite{li-etal-2023-api, yao2024taubench}, being time-consuming and labor-intensive. Alternatively, existing task-oriented dialogues share lots of same assumption and similarities with tool learning interactions, such the concept of domain (or services), intent, slots and corresponding values to help the users to complete the predefined tasks \citep{budzianowski-etal-2018-multiwoz, zhu-etal-2020-crosswoz}. Therefore, it is natural and straightforward to re-formulate task-oriented dialogues as a multi-turn interactions with tool and language feedback. To achieve this, we make several essential adjustments to align with the requirements of tool learning across the action, App, API and argument levels.


\begin{itemize}[leftmargin=*,itemsep=1pt] 
    \item \textbf{Actions.} The actions in TDD are typically defined in the format of \textit{intent-domain-slot-value} in lots of previous works \citep{kale-rastogi-2020-template, kwan-etal-2024-jotr-joint}. For example, "\textit{request-restaurant-name-?}" indicates that the system needs to request the name of the restaurant from the user. We save the first slot (a.k.a., request) since the other keys are all related to external tools/services, and then categorise them into \textit{non-tool} actions and \textit{tool-related} actions (shown in Table~\ref{tab:sys_actions}), considering lots of dialogue turns do not require tool calls. -- \textbf{Tool Awareness} \footnote{We note that our tool awareness is not simply binary classification (yes or no) but more fine-grained classification based on tool states.}
    
    \item \textbf{Apps.} We consider different domains/services as different Apps in a virtual mobile phone. Specifically, the schema of each App contains three key fields besides the API functions: 1) \textit{description}: which describes all tasks supported by the App in natural language; 2) \textit{base required arguments}: which provides information about base arguments required by all supported API in the App, such as user name and password; 3) \textit{APIs description}: using the name of API as key and store all necessary information about the APIs (see below). This kind of design enables more flexible and easy implementation by simply passing different fields to LLMs to decide which App or API to call. -- \textbf{Tool Selection}
    
    \item \textbf{APIs.} Following \citep{rastogi2020towards}, each API includes the \textit{name}, \textit{description}, a \textit{flag} \footnote{This is unique in SGD datasets since it requires the system to confirm before call the transactional API.} which indicates that the underlying API call is transactional (e.g, a booking or a purchase) as opposed to a search call, along with \textit{additional required} and \textit{optional arguments}. This setup closely resembles a function call in a programming language, particularly with regards to optional arguments. For instance, when renting a car, individuals may have varying preferences for the type of car they prefer. Taking these preferences into account enhances the tool's learning by making it more personalized and customized. The API can only be executed when all base required arguments and all additional required arguments are filled.  -- \textbf{Tool Selection}

    \item \textbf{Arguments.} Each argument is defined with the format: "\textit{name (type)}": "\textit{description}". For example, the argument start\_date in the `getcarsavailable' API is defined as "start\_date (date)": "the first date to start using the rental car, the format follows yyyy-mm-dd". We emphasize that the format of arguments is crucial when calling APIs, as they often expect specific structures or data types to function correctly. This is where LLMs can be particularly useful to interpret natural language instructions and convert them into the precise format required by the API. Using the same example above, the human may provides start date in a more casual format like "tomorrow" or "next Monday", LLMs need to translate that into the appropriate format (yyyy-mm-dd) for the API to understand (a.k.a, \textit{arguments formatting}). An example can be found in Fig~\ref{fig:app_api_schema}.  -- \textbf{Tool Execution}
    

\end{itemize}

\paragraph{Role Playing.} In order to provide customized experience and engaging users, we manually gather 50 roles (from movies or TV shows) along with their respective configurations, which include \textit{name}, \textit{gender}, \textit{MBTI type}, and \textit{behavioral attributes} following recent studies \citep{zhou2023characterglm}. These attributes encompass \textit{language features}, \textit{emotional expressions}, and \textit{interaction patterns}, leading to different conversation styles of dialogue assistant. Since all the previous operations do not change the content of utterances of the user and system except the progress annotations of each turn in multi-turn dialogues, we can directly prompt different LLMs to convert original system responses\footnote{We do not need to change user's utterance, and we find this does not affect the natural flow of the dialogues.} into role-consistent expression. This allows us to mimic how different individuals might convey the same results returned by APIs in distinct ways since response styles are orthogonal to the API results.



\subsection{Environment Set Up}

To mirror real-world agent-tool interactions, we need to carefully construct the tool environment and collect corresponding database. Firstly, we manually implement each App and API in python language by using the name of App as class name and each API within App as one function. We additionally add language descriptions for each App, API and corresponding arguments, and store them as special attribute of App class. There are other necessary functions and attributes to track the states of different App and APIs as the conversation goes on. Secondly, we sample every database lookup operation from the original dialogue datasets and store all unique returned results as the database for each App. For example, if the restaurant lookup operation returns several different candidates, we can store them together into the database. Afterwards, we can successfully build an virtual mobile environment -- \texttt{VirtualMobile} which enables interactive tool utilization and validation of correctness of created tools.


\begin{table}[t]
    \centering
    \begin{adjustbox}{max width=0.4 \textwidth}
    \begin{tabular}{l | cc}
    \toprule
    \textbf{Statistics} & \hspace{4mm} \textbf{Training} \hspace{2mm} & \hspace{2mm} \textbf{Evaluation} \hspace{4mm} \\
    \hline
    \# of Apps & 20 & 15 \\
    \# of APIs & 45 & 30 \\
    \# of Dialogues & 16,142 & 900 \\
    \# of Multiple Apps & 10,739 & 360 \\
    \# of Turns & 329,964  & 15,568  \\
    \# of Calls & 85,191 & 4,274 \\
    \# of Roles & 50 & 16 \\
    Avg. turns  & 20.4 & 17.3 \\
    Max. input arguments & 9 & 7 \\
    Avg. input arguments & 3.3 & 3.8 \\
    \bottomrule
    \end{tabular}
    \end{adjustbox}
    \caption{The data statistics of \texttt{DialogTool}. \# multiple apps means the number of dialogues where multiple apps are used.}
    \label{tab:data_stat}
\end{table}


\subsection{Data Analysis}

Table~\ref{tab:data_stat} presents the data statistics of our proposed \texttt{DialogTool}. The whole dataset contains 20 different Apps and 45 distinct APIs\footnote{The full list of Apps and APIs can be found in Appendix.}. Most Apps contain at least 2 APIs and 5 at maximum, covering lots of user needs in practice, such as booking a hotel/restaurant/flight, renting a car, finding near events and others. In general, \texttt{DialogTool} comprises approximately 16k dialogues and 33k turns, surpassing existing benchmarks like API-Bank \citep{li-etal-2023-api} by a significant magnitude. This scale enables comprehensive exploration of dialogue system capabilities across a vast corpus of interactions. Moreover, the dataset's average of around 4 input arguments per API during evaluation and over 16 turns per dialogue highlights the depth and complexity inherent in user-dialogue system interactions. This complexity is further underscored by the prevalence of multi-App dialogues, accounting for 50\% during training and 36\% during evaluation, showcasing the real-world challenge of orchestrating seamless interactions across multiple APIs.

\section{Experiments}
\label{sec:exp}

\subsection{Task Definition}

Given the dialogue context $c = \{u_1, s_1, ..., u_t\}$ and a virtual mobile environment $\mathcal{E} = \{App_1, App_2, ..., App_n\}$ where each App contains several APIs $\{p_i^1, ..., p_i^j\}$, and corresponding environmental state $e_{t-1}$ at current turn $t$, the dialogue agent either interact with the environment and then generate the final dialogue response $s_{t}$ according to updated state $e_t$ and previous context, or directly generate the response $s_{t}$ based on current state and context since it is not required to call the API in the environment.


\subsection{Set Up}
\label{setup}

Considering the complexity and interactivity of the whole life cycle of stateful tool use, we present more details about the evaluation for each dimension.

\paragraph{Tool Creation.} To assess LLMs' ability to develop new APIs, we provide them with complete information about the API, including its description, and all arguments (i.e, required and returned arguments) alongside with one demonstration. This ensures that the LLM understands the input, output, and purpose of the API before generating corresponding functions. Then we utilize all API calls in the test dialogues as test cases to evaluate whether or not created API (in the python function format) can successfully return the same results given same input arguments \footnote{The average (min) test cases for each API is 363 (93), and there are a total of 32 APIs that need to be created.}. Furthermore, this strict evaluation helps minimize the impact of code hallucination. We emphasize that this can be achieved on the fly during the conversation if existing toolsets are insufficient.

\paragraph{Tool Awareness.} Previous studies simply consider tool awareness as a binary classification problem such as using or does not use tools. However, it becomes inadequate in the context of stateful tool use since a user may inquiry about results of previous tool calls without necessitating a new tool invocation. Therefore, a more nuanced evaluation is needed to accurately reflect the complexities of stateful interactions. We consider more practical actions which support stateful multi-turn interactions as shown in Table~\ref{tab:sys_actions}, and we prompt the dialogue agent to select correct action $a$ from the given list based on current dialogue context $c$.

\paragraph{Tool Selection and Execution.} Once the previous determined action necessitates the API call, the dialogue agent needs to select the most appropriate API from the whole API list supported in the environments given the dialogue context $c$ and the environment $\mathcal{E}$, following the format of $\{t (k_1=v_1, ..., k_m=v_m)\}$. The $k$ and $v$ stands for the name and value of each argument of the API. We also consider hierarchical selection strategy which select appropriate App first and then select appropriate API, in order to better reflect the interactions of real-world applications. After it try to execute the tool at the environment, the environmental state will be updated to $e_t$ and then be used to generate the response. We calculate the accuracy at the API level and Argument level\footnote{If the user does not specify all necessary arguments in one turn, we let LLMs to replace the value with "?" for these missing arguments.} respectively. 

\paragraph{Response Generation.} According to the determined action, the dialogue agent generate the final response $s_t$ based on context $c$ and environmental state $e$, such as requiring more details about arguments and providing alternative suggestion regarding previous tool call results.

\paragraph{Role Play.} We additionally evaluate whether or not the dialogue agent can play different roles to transform the dialogue response $s_t$ into different styles $s_t^r$. Specifically, we randomly sample one role from predefined role list for one dialogue, the dialogue agent is tasked with generating role-consistent response. 



\subsection{Implementation Details}

\paragraph{Models.} We choose 13 distinct models whose size ranging from 6B to the 72B, aiming to provide comprehensive evaluation for current LLMs, following \citet{wang-etal-2024-appbench}. Specifically, we choose ChatGLM \citep{du2022glm} (\texttt{chatglm3-6b}), Qwen series \citep{qwen} (\texttt{Qwen1.5-7B/14B/72B-Chat}), Mistral \citep{jiang2023mistral} (\texttt{Mistral-7B-Instruct-v0.2}), LLaMa2 series \citep{touvron2023llama} (\texttt{Llama-2-7b/13b/70b-chat-hf}), and latest LLaMa3 series \citep{llama3modelcard} (\texttt{Meta-Llama-3-8B/70B-Instruct}) for open-source LLMs. Besides that, we also select latest GPT3.5 (\texttt{gpt-3.5-turbo}) and GPT-4o (\texttt{gpt-4o}) from OpenAI for closed-source LLMs. We set temperature and top p as 0.1 to reduce randomness. All experiments are run on NVIDIA A100 GPUs. 

\paragraph{Metrics.} For tool creation, we focus on the generated function can pass all test cases available in the \texttt{DialogTool}. The pass rates of each API function are then aggregated to determine overall performance (i.e., the ratio of passed test cases to total test cases), aligning with code tests\footnote{\url{https://leetcode.com/}}. For tool utilization, we adopt \textit{accuracy} to evaluate the performance following \cite{huang2024planning}. To assess the quality of generated responses, we employ well-established metrics such as BLEU and Rouge.L following previous studies \citep{li-etal-2023-api}. Furthermore, we employ GPT-4o to evaluate the consistency of roles depicted within the responses \citep{zhou2023characterglm}\footnote{We try other models such as Llama3.1-70B-Instruct and we do not observe significant differences.}. We also conduct a human evaluation to validate the alignment of our response evaluation setting with human judgements. We provide all details about the prompts and human study at the Appendix to ensure the reproductivity.


\begin{table*}[t]
    \setlength{\belowcaptionskip}{0pt}
    \centering
    \begin{adjustbox}{max width=0.85 \textwidth}
    \begin{tabular}{l| c| ccc | cccc}
    \toprule
    \multirow{2}{*}{\textbf{Models}} & \multirow{2}{*}{\textbf{Tool Creation}} & \multicolumn{3}{c|}{\textbf{Tool Utilization}} & \multicolumn{4}{|c}{\textbf{Role-consistent Responses}} \\    
    \cline{3-9} & & Awareness & Selection & Execution & BLEU & R.L & Role & Human \\
    \hline
    ChatGLM3-6B & 31.5 & 58.9 & 32.8 & 6.8 & 7.8 & 7.5 & 4.8 & 1.64 \\
    LLaMA2-7B & 33.2 & 63.5 & 27.4 & 7.0 & 6.8 & 5.7 & 6.2 & 1.25 \\
    QWen1.5-7B & 21.9 & \underline{68.9} & 54.7 & 11.3 & 8.0 & 7.4 & 7.0 & \underline{2.82} \\
    Mistral-7B & 11.4 & 42.5 & 51.8 & 22.6 & 8.0 & 7.1 & 6.7 & 2.28 \\
    LLaMA3-8B & \underline{62.2} & 46.3 & \underline{61.4} & \underline{45.6} & \underline{8.3} & \underline{7.7} & \underline{7.0} & 2.69 \\
    
    \hdashline
    
    LLaMA2-13B & \underline{48.8} &  47.1 & 51.1 & 11.7 & 7.7 & 6.4 & 6.5 & 2.17 \\
    Vicuna-13B & - & \underline{64.5} & \underline{62.9} & 12.3 & \underline{10.1} & \underline{11.5} & 6.0 & \underline{2.59} \\
    QWen1.5-14B & 27.9 & 51.7 & 55.6 & \underline{21.8} & 9.3 & 10.9 & \underline{7.5} & 2.44 \\

    \hdashline
    QWen1.5-72B & 49.7 & \textbf{75.5} & \underline{71.9} & 49.3 & \underline{10.8} & \textbf{15.3} & 7.4 & \underline{3.37} \\
    LLaMA2-70B & 23.0 & 34.7 & 57.8 & 32.6 & 8.5 & 10.7 & 6.2 & 2.56 \\
    LLaMA3-70B & \textbf{69.7} & 40.2 & 57.1 & \underline{68.1} & 9.0 & 11.3 & \underline{7.7} & 2.98 \\
    
    \hdashline
    
    GPT-3.5 & 63.3 & \underline{67.9} & 50.0 & 42.6 & 10.2 & 11.9 & 6.7 & 3.42 \\
    GPT-4o & \underline{66.7} & 63.5 & \textbf{77.8} & \textbf{68.7} & \textbf{11.4} &  \underline{14.5} & \textbf{8.3} & \textbf{3.56} \\
    \bottomrule
    \end{tabular}
    \end{adjustbox}
    \vspace{2mm}
    \caption{The main results of \texttt{DialogTool} at three stages: 1) Tool Creation; 2) Tool Utilization; 3) Role-consistent Response. \textbf{Bold} highlights the best score and \underline{underline} underscores the best score under the same model scale.}
    \label{tab:main_exp}
\end{table*}

\subsection{Main Results}

Table~\ref{tab:main_exp} shows the results of whole lifecycle of stateful tool use, several observations can be drawn as follows. 

\paragraph{Overall.} On the model side, GPT-4o outperforms other models in most cases, while QWen1.5-72B and LLaMA3-70B show competitive performance against GPT-4o. It is observed that the performance correlates positively with model size, particularly within the same model family. On the task side, no LLMs achieve an accuracy exceeding 80\% at the tool creation and utilization, and most LLMs performed poorly in tool creation and execution tasks compared to their performance in awareness and selection tasks, revealing the complexity and challenges of our dataset and environment. 

\paragraph{Tool Creation.} We find that LLaMA3 series model achieves exceptional performance at tool creation, such as LLaMA3-70B outforms GPT-4o and LLaMA3-8B is comparable with GPT-3.5. It can be attributed to additional code pre-training at the LLaMA3 models. In addition, Vicuna-13B can not pass the tool generation task and get no evaluation result, since the generated code has no indent and is not executable by python interpreter. Similar situation is observed on Mistral-7B, however, in some cases, Mistral-7B can still generate executable code, resulting in a low but non-zero pass rate.


\paragraph{Tool Utilization.} (1) \textbf{Awareness.} It is observed that the performance is not improved consistently as size increases, as validated by both QWen and LLaMa2. In addition, most of LLMs prefer not to use external tools (i.e., decide more non-tool actions: \textit{inform} or \textit{offer\_intent}) no matter small-sized models (Mistral-7B, LLaMA3-8B) or large-sized models (LLaMA2-70B). (2) \textbf{Selection and Execution.} A successful tool execution requires the correctness of API and all necessary arguments in the required format. Therefore, we can observe the performance of selection is better than execution in almost all LLMs, revealing the complexity of tool execution.

\paragraph{Role-consistent Response.} (1) \textbf{Response.} The larger the model size and the more accurate the tool utilization, the better the responses. This is reasonable since the results of tool utilization highly affect the quality of system response and the flow of conversation. For instance, LLaMA3-8B achieves better performance compared with other 7B models, and LLaMA3-70B further boost the performance due to increased size and more accurate tool utilization. (2) \textbf{Role Playing.} Larger models generally tend to deliver better performance, despite the gap in role-consistent scores across different LLMs being relatively small. (3) \textbf{Human Eva.} The trend is similar with what we observed during main experiments. The larger models tends to lead better performance, while GPT series models achieve best performance compared with other models.

\section{Analysis}

To offer a comprehensive evaluation of whole lifecycle of stateful tool use, we conduct error analyses for tool creation and tool utilization. In addition, we also explore the effects of different selection strategies \footnote{More analysis can be found in Appendix.}.

\paragraph{Tool Creation}

Figure~\ref{fig:appendix_tool_creation} shows the detailed tool creation performance of each API for each LLM. From the results, we can find that: \textit{Model size does not generally contribute to a higher performance.} We attribute this findings to the `hallucination' of code generation in tool creation task. For example, LLaMA2-70B tends to use \texttt{sqlite3} library that is not supported and required in the API description, while the smaller LLaMA2-7B and LLaMA2-13B does not have this issue. \textit{Complex tool is harder to be created than simple tool.} We observe a general lower performance across all LLMs on complex APIs like \texttt{gettraintickets} in the \texttt{Train} App which has 6 required arguments. In contrast, we find a general better performance on simple APIs like \texttt{schedulevisit} in the \texttt{Home} App that only has 2 required arguments. 

\begin{table}[!t]
    \setlength{\belowcaptionskip}{0pt}
    \centering
    \begin{adjustbox}{max width=0.48 \textwidth}
    \begin{tabular}{l| cc|ccc}
    \toprule
    \multirow{2}{*}{\textbf{Models}} & \multicolumn{2}{c|}{\textbf{Tool Awareness}}($\downarrow$) & \multicolumn{3}{c}{\textbf{Tool Selection}}($\downarrow$) \\ 
    \cline{2-6} & \textbf{T.} & \textbf{No T.} & \textbf{Rec.} &\textbf{ API.} & \textbf{Unn.} \\
    \hline
    QWen1.5-7B & \textbf{2.5} & 28.6 & 76.8 & 0.8 & 28.9 \\
    LLaMA3-8B & 20.8 & 32.9 & 64.6 & 1.2 & 4.0 \\
    
    \hdashline
    
    LLaMA2-13B & 19.6 & 33.3 & 59.9 & 4.1 & 13.1 \\
    QWen1.5-14B & 21.8 & 26.5 & 46.8 & 11.3 & 3.7 \\

    \hdashline
    QWen1.5-72B & 7.3 & 17.2 & 84.2 & 0.2 & 11.5 \\
    LLaMA3-70B & 39.2 & 20.6 & 82.6 & \textbf{0.0} & 0.4 \\
    
    \hdashline
    
    GPT-3.5 & 4.8 & 27.3 & 68.6 & 0.1 & 27.1 \\
    GPT-4o & 19.9 & \textbf{16.7} & 91.4 & 0.9 & \textbf{0.3} \\
    \bottomrule
    \end{tabular}
    \end{adjustbox}
    \caption{The error analysis of tool awareness and tool selection. \textbf{T.}: the rates of cases that LLMs should use tool but the action does not invoke it (false negative); \textbf{No T.}: The rates of cases that LLMs should not use tool but the action invoke tool call or invoke other types of \textit{non-tool} actions (false positive). \textbf{Rec.} \textbf{API.} and \textbf{Unn.} stand for the recall, API parsing error and unnecessary tool calls.}
    \label{tab:analysis}
\end{table}

\begin{figure}
    \centering
    \includegraphics[width=0.49\textwidth]{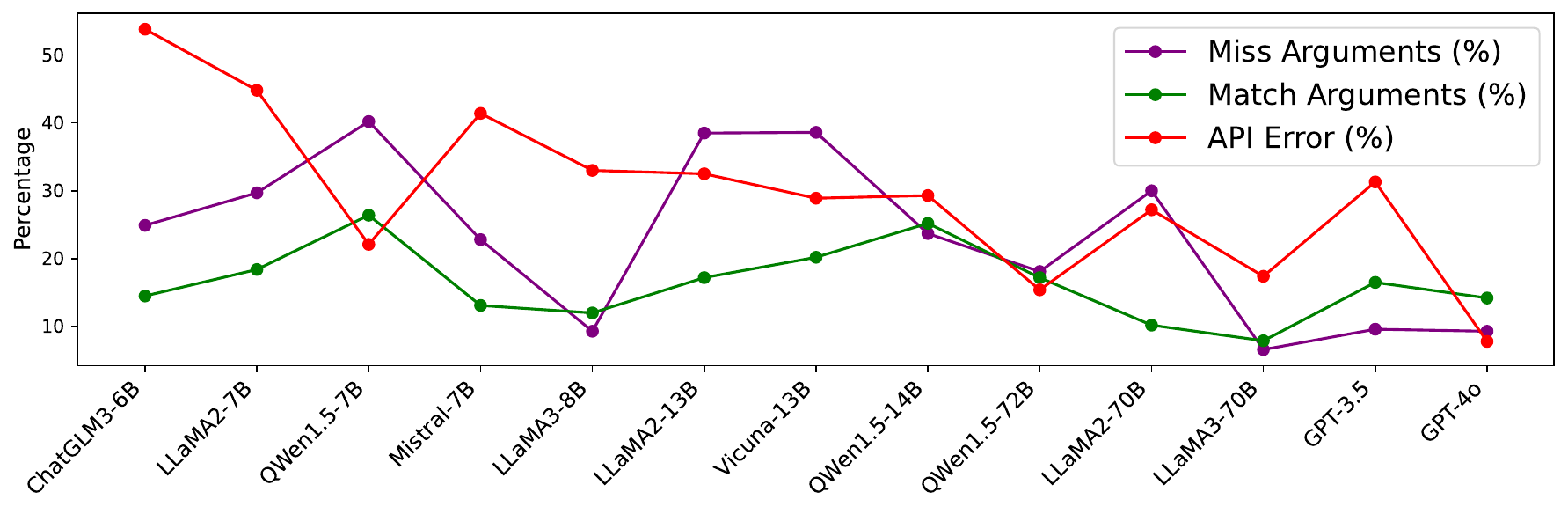}
    \caption{The three primary errors at tool execution.}
    \label{fig:error_tool_exe}
\end{figure}

\begin{figure}[t]
\centering
\begin{subfigure}{.35\textwidth}
  \centering
  \includegraphics[trim={1cm 2cm 0cm 1cm}, clip, scale=1.0, width=.7\textwidth]{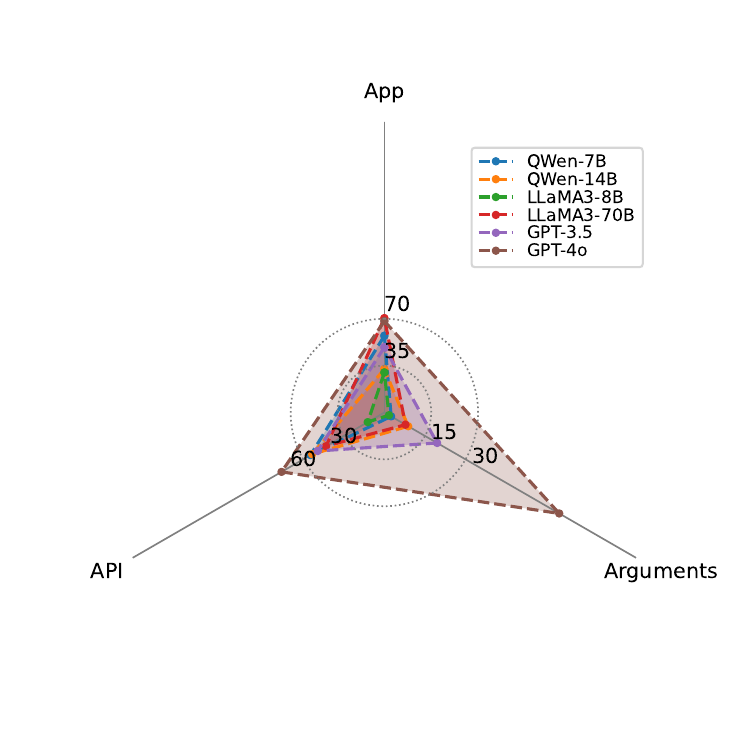}
  \caption{Tool Selection under \textit{Flat} Setting}
  \label{fig:sub1}
\end{subfigure}%
\hspace{0.01\textwidth} 
\begin{subfigure}{.35\textwidth}
  \centering
  \includegraphics[trim={0cm 1cm 1cm 1cm}, clip, scale=1.0, width=.7\textwidth]{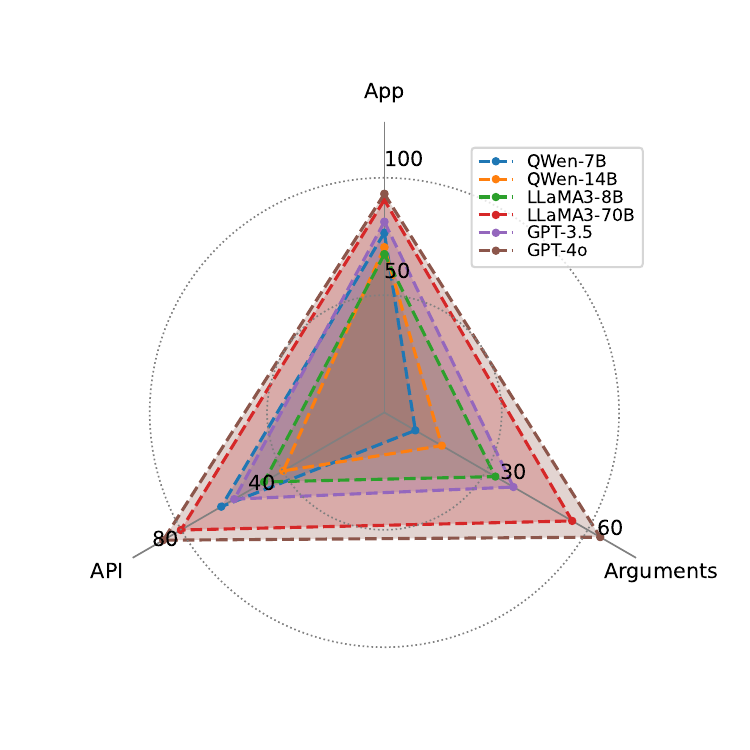}
  \caption{Tool Selection under \textit{Hierarchical} Setting}
  \label{fig:sub2}
\end{subfigure}
\caption{The performance of two prompting strategies.}
\label{fig:flat_and_hier}
\end{figure}

\paragraph{Tool Utilization}

i) \textit{Tool Awareness and Selection.} Table~\ref{tab:analysis} shows the error analyses. On the one hand, we can find LLMs struggle to select the timing to use tool, resulting in a high rate of false negatives or false positives. Furthermore, the performance gap across different LLMs mainly comes from the FN part especially for powerful LLMs since they tend to not use external tools. On the other hand, the recall is usually higher than acc regardless of the type of LLM. This indicates that once LLMs know when to call tools, they have a good chance of choosing the correct one. Moreover, we can attribute the relatively lower accuracy of GPT-3.5 to the 27.1\% of its tool calls being unnecessary; ii) \textit{Tool execution.} Figure~\ref{fig:error_tool_exe} shows the three major errors. It is observed that most of error cases comes from missing arguments instead of arguments matching (i.e, name and value matching). In addition, API error remains a significant challenge, highlighting the difficulties associated with  error propagation in the task.

\paragraph{Hierarchical Tools}
\label{hier_tool_select}

We examine the performance of LLMs in selecting APIs when the App is not known in advance. Therefore, we explore two scenarios: 1) \textit{Flat}: select APP and API together; and 2) \textit{Hierarchical}: select APP first and then API. Figure~\ref{fig:flat_and_hier} illustrates the results. \textit{Generally, we can find that models likely achieves much higher performance under hierarchical setting instead of flat one.} We emphasize this not only support the effectiveness of our introduced hierarchical structure, but also more aligns with the cases in practice since it is usually impossible to access all APIs inside each App for the dialogue agent. \textit{Furthermore, it is noteworthy that the performance of LLMs on arguments level is worst compared with App and API selection in both settings}, since it requires LLMs to recognize and extract the values for each arguments from multi-turn dialogue interaction, and then format them in the required format. This also aligns with our main experimental results.

\begin{table}[t]
\centering
\begin{adjustbox}{max width=0.45 \textwidth}
\begin{tabular}{c|c|c|c}
\hline
\textbf{Turns} & \textbf{Tool Awareness} & \textbf{Tool Selection} & \textbf{Tool Execution} \\
\hline
10 & 78.9 & 85.7 & 75.2 \\
20 & 59.7 & 69.5 & 51.2 \\
30 & 56.7 & 60.1 & 38.5 \\
40 & 52.3 & 54.4 & 35.3 \\
\hline
\end{tabular}
\end{adjustbox}
\caption{Performance across different number of turns for Tool Awareness, Selection, and Execution.}
\label{tab:tool_performance_turns}
\end{table}

\paragraph{Effects of Number of Turns.} We additionally conduct performance of number of turns based on GPT-4o model. Specifically, there are many turns in the multi-turn dialogues ranging from 4 to more than 40 turns. Table~\ref{tab:tool_performance_turns} shows the final results. It is found that as the number of turns increases, the performance drops accordingly, especially for the tool execution.





\section{Conclusion}

We introduce \texttt{DialogTool}, which is the first multi-turn interactive benchmark considering the whole life cycle of tool utilization of dialogue agent, spanning across three distinct stages, six different tasks. Furthermore, we build \texttt{VirtualMobile}, an virtual mobile evaluation environment to simulate API calls and return corresponding results. We hope we benchmark and environment will provide a comprehensive platform for evaluating and advancing dialogue agents' tool utilization capabilities, fostering future research in multi-turn interactions.


\section*{Limitation}

Our study comes with two minor limitations. On the one side, each App contains a limited number of APIs, focused primarily on key functions. We start with the most commonly used App and APIs in the real-world and target the interactions between the user, dialogue agent and the environment. As revealed by recent studies \cite{yao2024taubench}, it is hard for existing advanced LLMs to successfully complete the task over long horizons even considering only two simple situations such as retail and airline.

On the other side, we do not consider the existing agent framework since we mainly focus on the base capabilities of various LLMs on this new benchmark. We anticipate that introducing additional reflection or a carefully designed agent workflow may further boost the performance on this setting.


\section*{Ethical Considerations}

In conducting our research, we have thoroughly reviewed and ensured compliance with ethical standards. Our study utilizes existing datasets, which have been publicly available and previously vetted for ethical use. These datasets have been carefully selected to avoid any form of offensive or biased content. Therefore, we consider that our research does not present any ethical issues. The data used is ethically sourced, the analysis is unbiased, and all procedures align with established ethical guidelines.

\section*{Acknowledgments}

This work was partially supported by Hong Kong RGC GRF No. 14206324, CUHK direct grant No. 4055209, and CUHK Knowledge Transfer Project Fund No. KPF23GWP20.

\bibliography{custom}

\appendix

\clearpage

\section*{Appendix}
\label{sec:appendix}

\section{Related Work}

\paragraph{Dialogue System.} On the one hand, task-oriented dialogue systems typically comprise four components Natural Language Understanding (NLU) \citep{goo2018slot}, Dialogue State Tracking (DST) \citep{wang-etal-2021-fast}, Dialogue Policy Learning (DPL) \citep{wang2022integrating, takanobu-etal-2019-guided}, and Natural Language Generation (NLG) \citep{peng2020few}, or adopt end-to-end framework \citep{qin-etal-2023-end} to complete user goal through complex multi-turn interaction with various services in real-world. On the other hand, open-domain dialogue systems mostly follow retrieval-augmented generation framework \citep{lewis2021retrievalaugmented, wang-etal-2023-large}  to retrieve different external knowledge such as persona \citep{xu-etal-2022-long} and document \citep{dinan2018wizard}, aiming to provide more personalized and informative responses.

\paragraph{Language Agent.} To evaluate the effectiveness of LLMs as agents, many prior works have proposed various evaluation benchmarks \citep{liu2023agentbench, ma2024agentboard, sumers2023cognitive}. For instance, VirtualHome \citep{VirtualHome} functions as a simulation platform for typical household activities, while ScienceWorld \citep{wang-etal-2022-scienceworld} assesses agents' scientific reasoning abilities within an interactive text environment. Distinguishing from these benchmarks, we target to build a virtual mobile environment where the LLMs need to call different APIs in various APPs in order to complete the complex goal of users via multi-turn dialogue interaction. Furthermore, our proposed benchmark can extend to multi-agent scenarios by conceptualizing large language models (LLMs) as operating systems (OS) and various applications (APPS) as distinct agents within the system \citep{ge2023llm}. Besides that, assign agents some pre-defined roles has been proved an effective way to engaging user, resulting in more longer interaction time, such as character.ai, and also elicit some reasoning and role-specific capabilities of LLMs \citep{zhou2023characterglm}. For example, RoleLLM \citep{wang2023rolellm} and CharacterEval \citep{tu2024charactereval} evaluate the LLMs to generate role-consistent responses according to given background config of the role. On the contrast, we mainly focus on the language style of different roles \citep{zhou2023characterglm}, leading to more personalized and acceptable response for users.

\section{Dataset Details}

\subsection{APP and API List}
\label{sec:app_api_list}

Table~\ref{tab:app_api_list} show the full list of supported APP and API.

\begin{table*}[]
    \centering
    \begin{tabular}{cl}
    \toprule
        \textbf{APP} & \textbf{API} \\ \midrule
        Banks & CheckBalance, TransferMoney \\ 
        Buses & BuyBusTicket, FindBus \\ 
        Events & FindEvents, GetEventDates, BuyEventTickets \\ 
        Flights & SearchRoundtripFlights, ReserveRoundtripFlights, SearchOnewayFlight, ReserveOnewayFlight \\ 
        Homes & FindHomeByArea, ScheduleVisit, FindApartment \\ 
        Hotels & ReserveHotel, SearchHouse, BookHouse, SearchHotel \\ 
        Media & FindMovies, PlayMovie, RentMovie \\ 
        Movies & FindMovies, BuyMovieTickets, GetTimesForMovie \\ 
        Music & PlayMedia, LookupSong, LookupMusic, PlaySong \\ 
        RentalCars & GetCarsAvailable, ReserveCar \\ 
        Restaurants & ReserveRestaurant, FindRestaurants \\ 
        RideSharing & GetRide \\ 
        Services & FindProvider, BookAppointment \\ 
        Travel & FindAttractions \\ 
        Weather & GetWeather \\ 
        Calendar & GetEvents, AddEvent, GetAvailableTime \\ 
        Alarm & GetAlarms, AddAlarm \\ 
        Messaging & ShareLocation \\ 
        Payment & MakePayment, RequestPayment \\ 
        Trains & FindTrains, GetTrainTickets \\ 
    \bottomrule
    \end{tabular}
    \caption{APP and API list.}
    \label{tab:app_api_list}
\end{table*}

\subsection{DialogTool Dataset}

Table~\ref{tab:sys_actions} show the full list of all actions in the \texttt{DialogTool}.

\begin{table*}[!h]
\setlength{\belowcaptionskip}{0pt}
    \centering
    \begin{adjustbox}{max width=1.0 \textwidth}
    \begin{tabular}{llc}
    \toprule
    \textbf{Name} & \textbf{Desc} & \textbf{Type} \\
    \hline
    Request & Request the value of an argument from the user & 
    \textit{tool-related} \\
    
    Confirm & Confirm the value of all arguments before making a transactional API call & \textit{tool-related} \\

    Inform\_Count & Inform the number of iterms found that satify user's request & \textit{tool-related} \\

    Notify\_Success & Inform the user that their request was successful & \textit{tool-related} \\

    Notify\_Failure & Inform the user that their request failed & \textit{tool-related} \\

    Inform & Inform the value for an argument to the user  & \textit{non-tool} \\
    
    Offer\_Intent & Offer a new intent to the user. Eg, "Would you like to reserve a table?" & \textit{non-tool} \\
    
    Req\_more & Asking the user if they need anything else & \textit{non-tool} \\
    
    Goodbye & End the dialogue & \textit{non-tool} \\
    
    \bottomrule
    \end{tabular}
    \end{adjustbox}
    \caption{Pre-defined Actions in SGD dataset. The name and descriptions are copied from SGD dataset \cite{rastogi2020towards}.}
    \label{tab:sys_actions}
\end{table*}

\subsection{Schema of App and API}
\captionsetup{type=figure}
\begin{lstlisting}
"Rents": {
  "desc": "a leading global provider of car rental solutions",
  "base_required_arguments": {},
  "APIs": {
    "getcarsavailable": {
      "desc": "discover cars available for rent in a certain location and period",
      "is_transactional": "False",
      "additional_required_arguments": {
        "city (str)": "city where you want to rent the car",
        "start_date (date)": "the first date to start using the rental car, the format follows yyyy-mm-dd.",
        "pickup_time (time)": "time for the pick-up, the format follows hh:mm",
        "end_date (date)": "the date to return the car, the format follows yyyy-mm-dd"
      },
      "optional_arguments": {
        "car_type (str)": "type of the car, value can only be one of follows: Hatchback, Sedan or SUV"
      },
      "result_arguments": {
        "car_type (str)": "type of the car, value can only be one of follows: Hatchback, Sedan or SUV",
        "car_name (str)": "car model",
        "pickup_location (str)": "place to pick up the car",
        "start_date (date)": "the first date to start using the rental car, the format follows yyyy-mm-dd",
        "pickup_time (time)": "time for the pick-up, the format follows hh:mm",
        "city (str)": "city where you want to rent the car",
        "end_date (date)": "the date to return the car, the format follows yyyy-mm-dd",
        "price_per_day (int)": "the cost for renting the car per day"
      }
    },
    "reservecar": {
      "desc": "make a rental car reservation",
      "is_transactional": "True",
      "additional_required_arguments": {
        "pickup_location (str)": "place to pick up the car",
        "start_date (date)": "the first date to start using the rental car, the format follows yyyy-mm-dd",
        "pickup_time (time)": "time for the pick-up, the format follows hh:mm",
        "end_date (date)": "the date to return the car, the format follows yyyy-mm-dd",
        "car_type (str)": "type of the car, value can only be one of follows: Hatchback, Sedan or SUV",
        "add_insurance (bool)": "whether to purchase insurance, True or False"
      },
      "optional_arguments": {},
      "result_arguments": {
        "car_type": "type of the car, value can only be one of follows: Hatchback, Sedan or SUV",
        "car_name": "car model",
        "pickup_location": "place to pick up the car",
        "start_date": "the first date to start using the rental car",
        "pickup_time": "time for the pick-up",
        "end_date": "the date to return the car",
        "price_per_day": "the cost for renting the car per day",
        "add_insurance": "whether to purchase insurance"
      }
    },
    "getride": {
      "desc": "book a cab for any destination, number of seats and ride type",
      "is_transactional": "True",
      "additional_required_arguments": {
        "destination (str)": "destination address or location for cab",
        "number_of_seats (int)": "number of seats to reserve in the cab",
        "ride_type (str)": "type of cab ride"
      },
      "optional_arguments": {},
      "result_arguments": {
        "destination": "destination address or location for cab",
        "ride_type": "type of cab ride, value can only be one of follows: Pool, Regular or Luxury",
        "ride_fare": "total fare for cab ride",
        "wait_time": "expected waiting time for pick-up by cab",
        "number_of_seats": "number of seats to reserve in the cab"
      }
    }
  }
}

\end{lstlisting}
\captionof{figure}{This is a sample JSON configuration of \texttt{Rents} App which contains 3 distinct APIs. We also provide name, format and possible values for categorical arguments. In this app, the base required arguments are empty.}
\label{fig:app_api_schema}

\section{Prompt Details}
\label{appendix:prompts}
\begin{tcolorbox}[breakable]
You are a helpful assistant and you are good at Python.

Given the description, required arguments, optional required arguments and returned arguments of an APIs, generate a executable python code which implements the API.

Here is an example:

API: 

\begin{lstlisting}
getcarsavailable: {
    "desc": "discover cars available for rent in a certain location and period",
    "is_transactional": False,
    "additional_required_arguments": {
        "city (str)": "city where you want to rent the car",
        "start_date (date)": "the first date to start using the rental car, the format follows yyyy-mm-dd.",
        "pickup_time (time)": "time for the pick-up, the format follows hh:mm",
        "end_date (date)": "the date to return the car, the format follows yyyy-mm-dd"
    },
    "optional_arguments": {
        "car_type (str)": "type of the car"
    },
    "result_arguments": {
        "car_type (str)": "type of the car",
        "car_name (str)": "car model",
        "pickup\_location (str)": "place to pick up the car",
        "start_date (date)": "the first date to start using the rental car, the format follows yyyy-mm-dd",
        "pickup_time (time)": "time for the pick-up, the format follows hh:mm",
        "city (str)": "city where you want to rent the car",
        "end_date (date)": "the date to return the car, the format follows yyyy-mm-dd",
        "price_per_day (int)": "the cost for renting the car per day"
    }
}
\end{lstlisting}
Python:

\begin{lstlisting}
def getcarsavailable(self, city, start_date, end_date, pickup_time, car_type=""):
    print("This is api [getcarsavailable] in [Rents] app")
    results = []
    for db_sample in self.db["getcarsavailable"]:
        if db_sample["city"] in city and db_sample["start_date"] == start_date and db_sample["end_date"] == end_date and \
            db_sample["pickup_time"] == pickup_time:
            if len(car_type) > 0 and db_sample["car_type"] == car_type:
                results.append(db_sample)
            elif len(car_type) == 0 or car_type is None:
                results.append(db_sample)
    return results
\end{lstlisting}

API: \textcolor{red}{\{api\_desc\}} 

Python:

\end{tcolorbox}
\caption{The prompt used to prompt LLM to create tool in python code.}
\label{fig:tool_creation}

\begin{tcolorbox}[breakable]

Given a dialogue between user and dialogue system, and a role config for dialogue system, please assign a consistency score according to all utterances by the dialogue system. The consistency of a role refers to the character's actions, dialogues, and decisions with their defined traits and background. The criteria for measuring "Consistency" are detailed in the following dimensions:

1. Behavioral Consistency: Evaluate whether the character's behavior aligns with their described personality and background across these aspects:

- Personality Display: Does the character exhibit personality traits in interactions that match their predefined descriptions? For example, if a character is described as brave, they should exhibit bravery in the face of danger.

- Background Response: Does the character's behavior in specific situations reflect their background knowledge and experiences? For instance, a character who was once stranded on a deserted island might display enhanced survival skills in similar settings.

- Emotional Consistency: Do the character's emotional responses align with the situation and their personal history?

2. Dialogue Consistency: Assess if the character's dialogue reflects their personality traits and background story:

- Language Style: Does the character's use of language suit their cultural and educational background?

- Relevance to Theme: Are the contents of the dialogue relevant to the character's life experiences and current situation?

- Emotional Expression: Does the character's emotional expression in dialogue match the personality described?

3. Decision-Making Consistency: Evaluate whether the character's decisions align with their goals and role setting:

- Goal Orientation: Do the decisions help the character achieve their set objectives?

- Background Logic: Do the decisions take into account the character's personal and societal background?

- Situational Appropriateness: Are the decisions reasonable and effective within specific scenarios?

Your output should range from 0 to 10, where 0 represents complete inconsistency and 10 represents perfect consistency. You only need to generate one score considering above factors without generating other information.

\end{tcolorbox}

\caption{The prompt used to prompt LLM to assign the role consistency score.}
\label{fig:role_consistency_score}

\begin{tcolorbox}[breakable]

Please determine which action of system should be invoked to generate the next response. Note some actions do not require the involvement of functional APIs. 

Here are all pre-defined actions starting from action\_name followed by descriptions: \textcolor{red}{\{sys\_actions\}}

You should output it in the format of [action\_name(explanations)]. Your output should start with a square bracket '[' and end with a square bracket ']'. Do not output any other explanation or prompt. The action\_name can only be one of pre-defined actions. You only need to choose one action according to what user needs.

\end{tcolorbox}

\caption{The prompt used to prompt LLM to generate an action.}
\label{fig:action_prompt}

\begin{tcolorbox}[breakable]
Given the API description and the existing dialogue history, please generate one API request that should be invoked to complete the user's current query, and output it in the format of [api\_name(\#argument\_1='value of argument\_1', \#argument\_2='value of argument\_2', ...)]. 

Here is the description of all APIs: \textcolor{red}{\{api\_desc\}}

Here is the current date: \textcolor{red}{\{date\}}, make sure the values of all date related arguments is based on current date. 

The api\_name can only be one of pre-defined apis. You need to list all additional\_required\_arguments in the corresponding API, and optional\_arguments when they are provided in the dialogue. 

You should replace the value with the actual value in the dialogue context and attention on the format requirements of each argument. You can use "?" to replace the value when you can not infer it via current context. 
    
Your output should start with a square bracket '[' and end with a square bracket ']'. Do not output any other unrelated explanation or tokens outside of [].

\end{tcolorbox}
\caption{The prompt used to prompt LLM to generate API and all related arguments appeared in the dialogue in the required format.}
\label{fig:api_prompt}

\begin{tcolorbox}[breakable]

Your task is to determine the required App according the description of each App and the last user turn in the dialogue. 

Here is the information about all accessible Apps: \textcolor{red}{\{api\_desc\}}
Your output should follow the format [app1, app2, ...]. You only need to output one App.

\end{tcolorbox}
\caption{The prompt used to prompt LLM to decide App first under the \textit{hierarchical} setting.}
\label{fig:analyze_prompt_hierarchy_step_1}

\begin{tcolorbox}[breakable]

Given the API description and the existing dialogue history, please generate one API request that should be invoked to complete the user's current query, and output it in the format of 
[api\_name(\#argument\_1='value of argument\_1', \#argument\_2='value of argument\_2', ...)]. The api\_name can only be one of pre-defined apis. You need to list all additional\_required\_arguments in the corresponding API, and optional\_arguments when they are provided in the dialogue.
You should replace the value with the actual value in the dialogue context and attention on the format requirements of each argument. You can use "?" to replace the value when you can not infer it via current context. 
Your output should start with a square bracket '[' and end with a square bracket ']'. Do not output any other unrelated explanation or tokens outside of []. 

Here is the current date: \textcolor{red}{\{date\}}, make sure the values of all date related arguments is based on current date. 

Here is the description of all APIs: \textcolor{red}{\{api\_desc\}}

\end{tcolorbox}
\caption{The prompt used to prompt LLM to decide API and corresponding arguments after decided App under the \textit{hierarchical} setting.}
\label{fig:analyze_prompt_hierarchy_step_2}

\begin{tcolorbox}[breakable]

Given the API description and the existing dialogue history, please generate one API request that should be invoked to complete the user's current query, and output it in the format of 
[app\_name: api\_name(\#argument\_1='value of argument\_1', \#argument\_2='value of argument\_2', ...)]. The app\_name and api\_name can only be one of pre-defined apps and one of pre-defined apis in the app. You need to list all additional\_required\_arguments in the corresponding API, and optional\_arguments when they are provided in the dialogue. 
You should replace the value with the actual value in the dialogue context and attention on the format requirements of each argument. You can use "?" to replace the value when you can not infer it via current context. 
Your output should start with a square bracket '[' and end with a square bracket ']'. Do not output any other unrelated explanation or tokens outside of []. 

Here is the current date: \textcolor{red}{\{date\}}, make sure the values of all date related arguments is based on current date. 

Here is the description of all Apps: \textcolor{red}{\{app\_api\_desc\}}
    
\end{tcolorbox}
\caption{The prompt used to prompt LLM to decide App, API and all related arguments at the same time under the \textit{flat} setting.}
\label{fig:analyze_prompt_flat}

\section{Analysis}
\label{appendix:ana}

\begin{figure*}[t]
    \centering
    \begin{subfigure}[b]{0.3\textwidth}
        \centering
        \includegraphics[width=\textwidth]{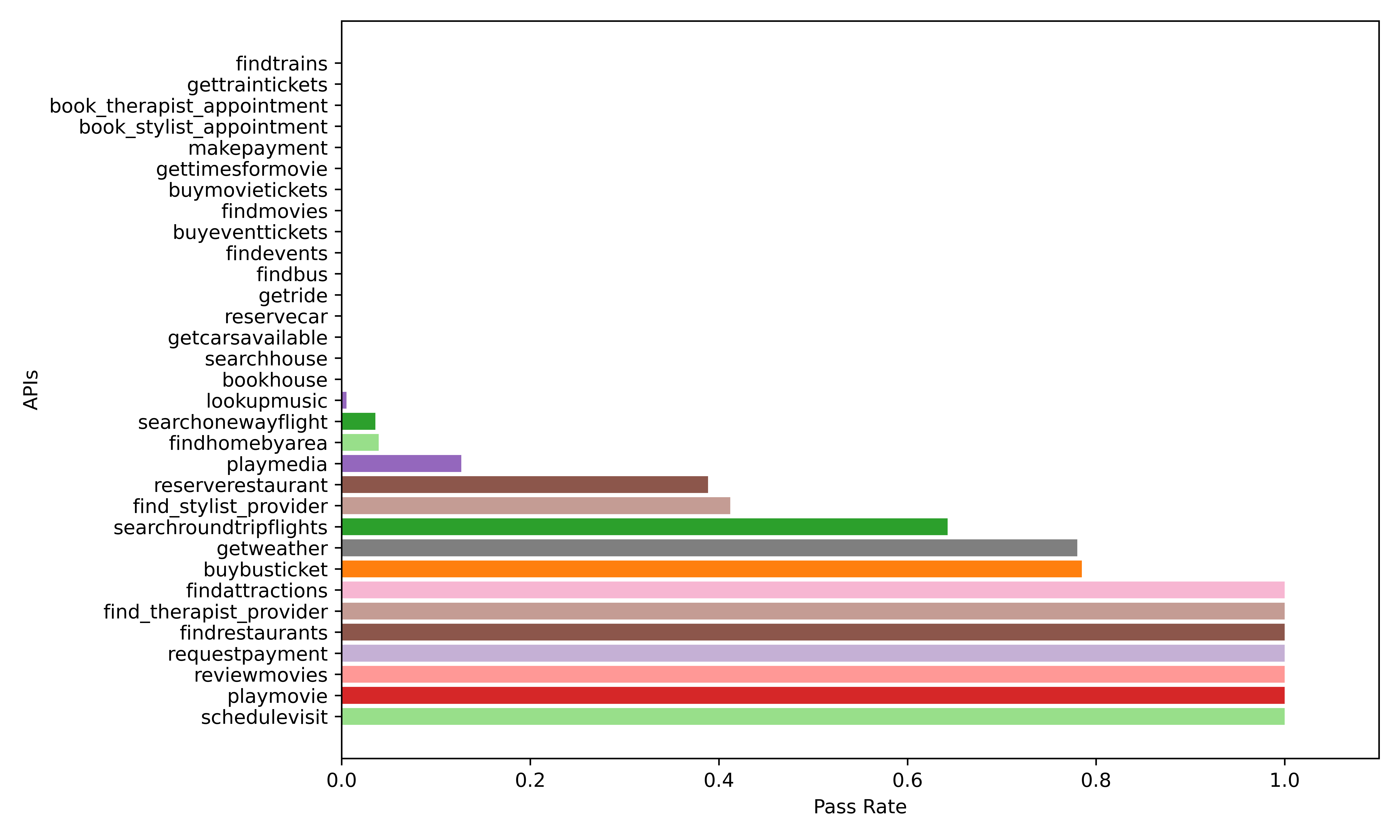}
        \caption{LLaMA2-7B}
        \label{fig:tc-llama2-7b}
    \end{subfigure}
    \hfill
    \begin{subfigure}[b]{0.3\textwidth}
        \centering
        \includegraphics[width=\textwidth]{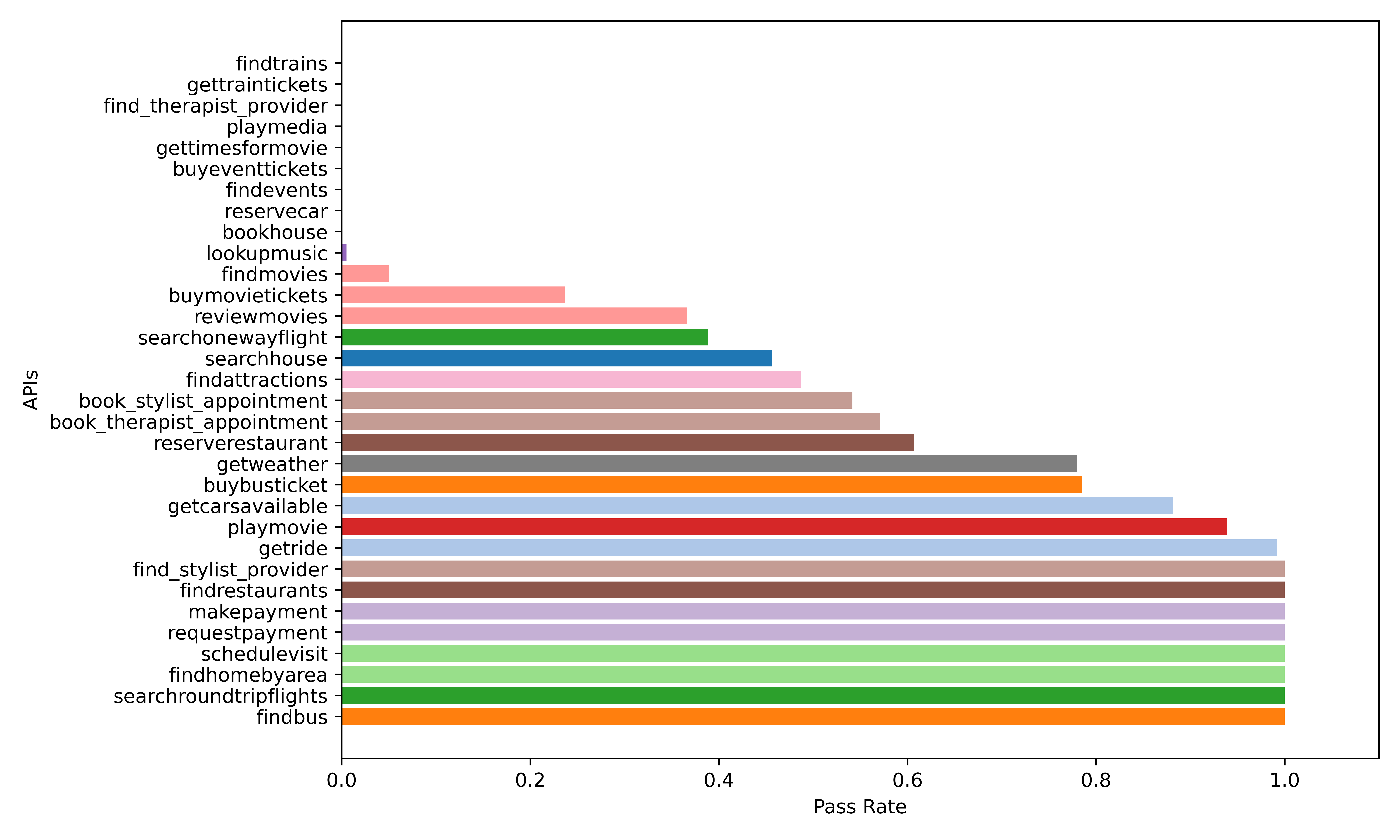}
        \caption{LLaMA2-13B}
        \label{fig:tc-llama2-13b}
    \end{subfigure}
    \hfill
    \begin{subfigure}[b]{0.3\textwidth}
        \centering
        \includegraphics[width=\textwidth]{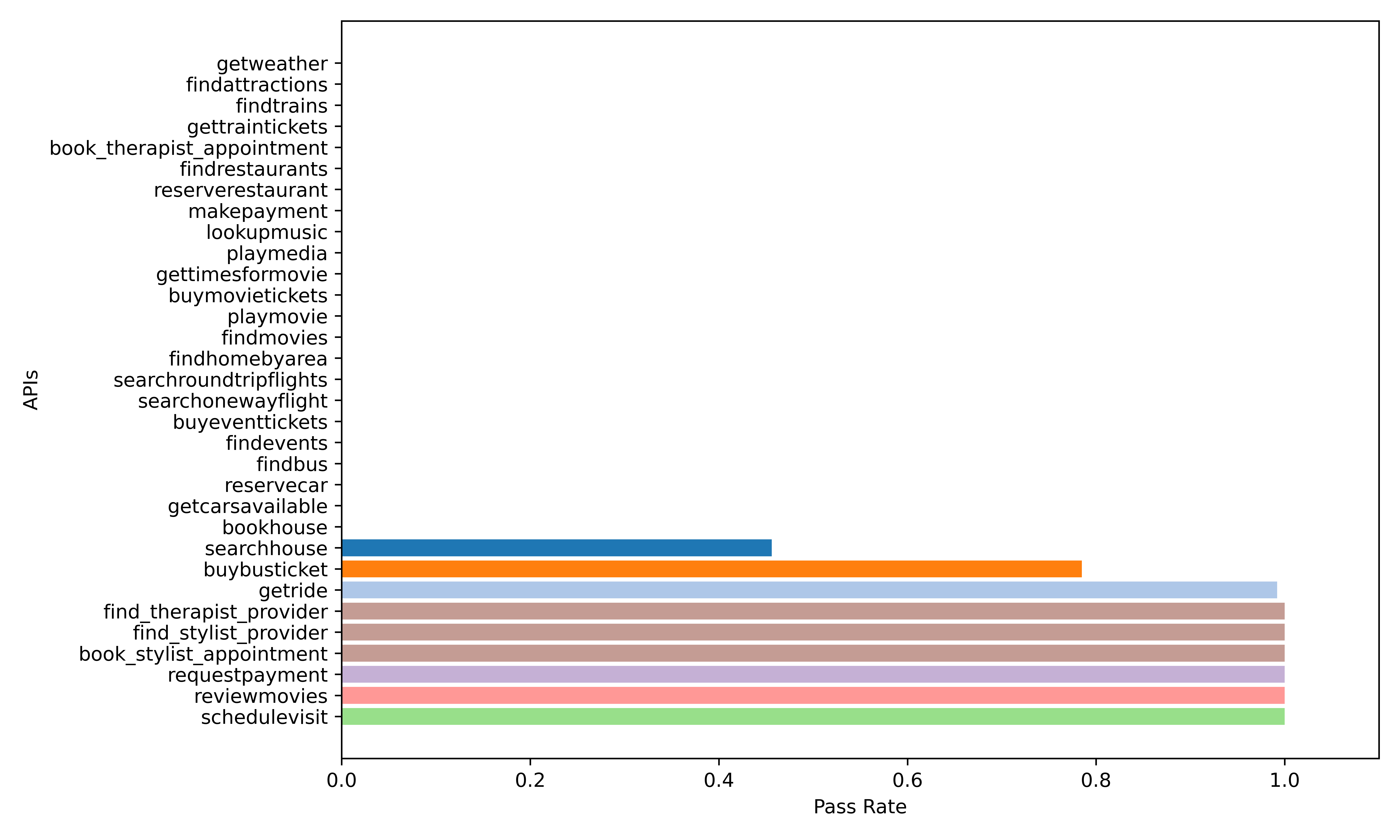}
        \caption{LLaMA2-70B}
        \label{fig:tc-llama2-70b}
    \end{subfigure}

    \vskip\baselineskip

    \begin{subfigure}[b]{0.3\textwidth}
        \centering
        \includegraphics[width=\textwidth]{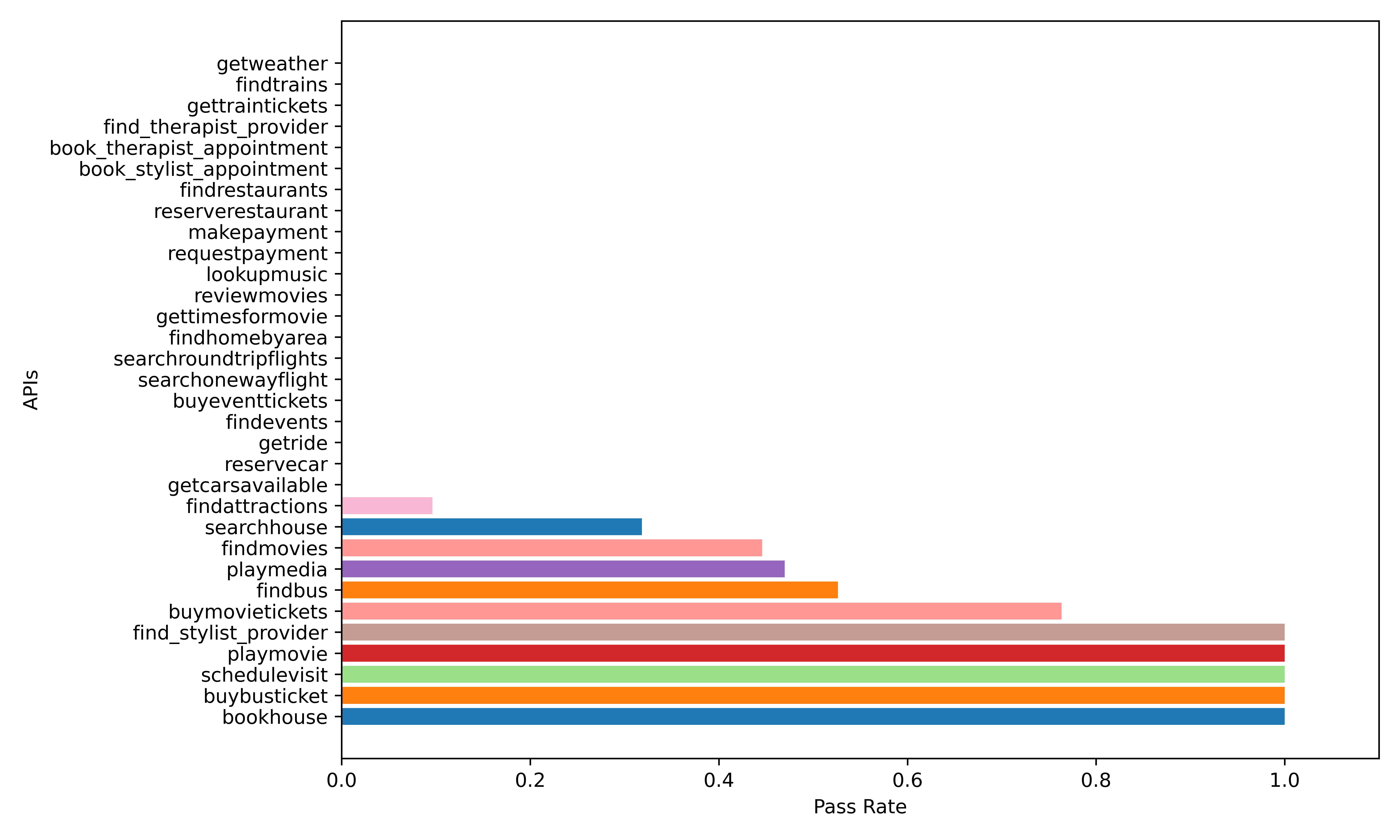}
        \caption{QWen1.5-7B}
        \label{fig:tc-qwen1.5-7b}
    \end{subfigure}
    \hfill
    \begin{subfigure}[b]{0.3\textwidth}
        \centering
        \includegraphics[width=\textwidth]{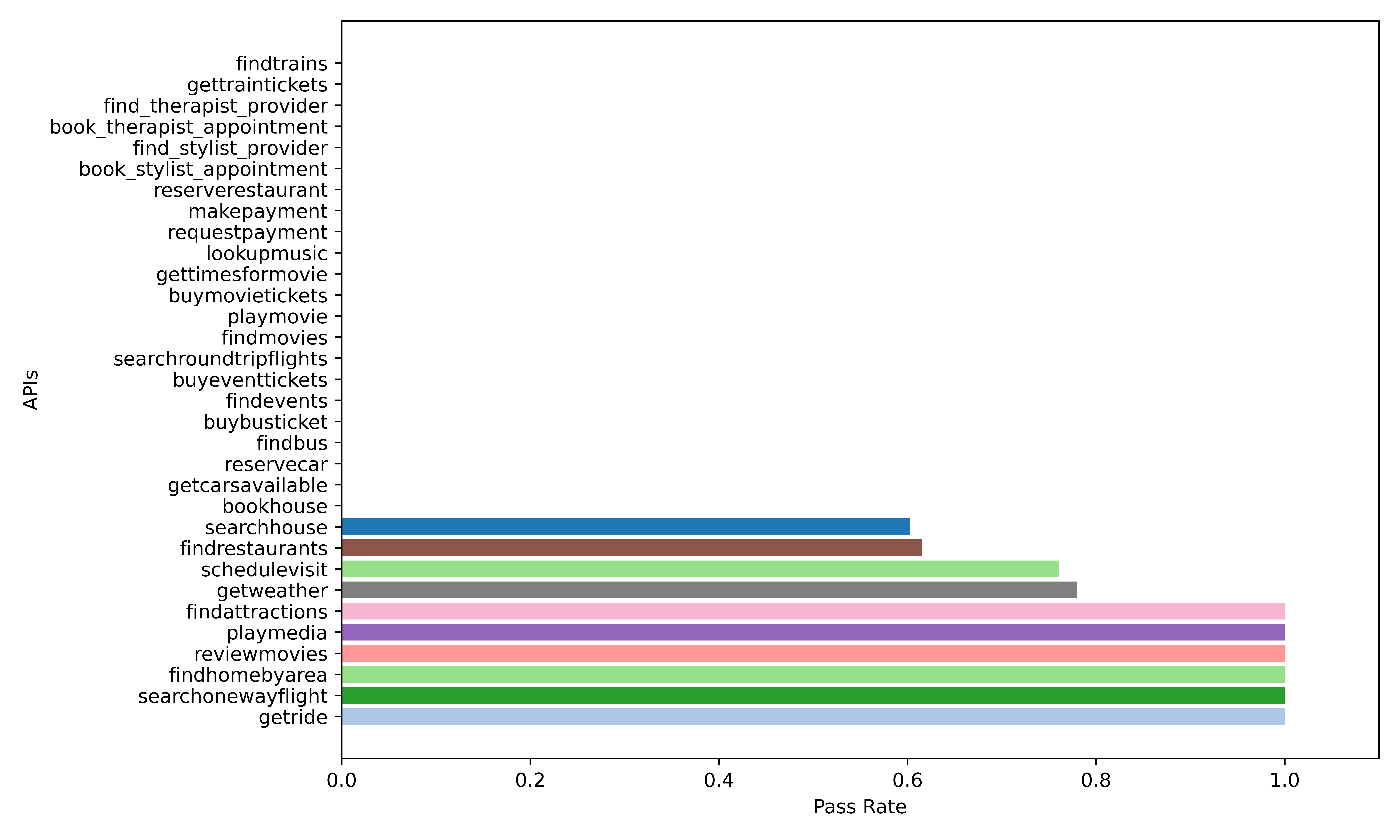}
        \caption{QWen1.5-14B}
        \label{fig:tc-qwen1.5-14b}
    \end{subfigure}
    \hfill
    \begin{subfigure}[b]{0.3\textwidth}
        \centering
        \includegraphics[width=\textwidth]{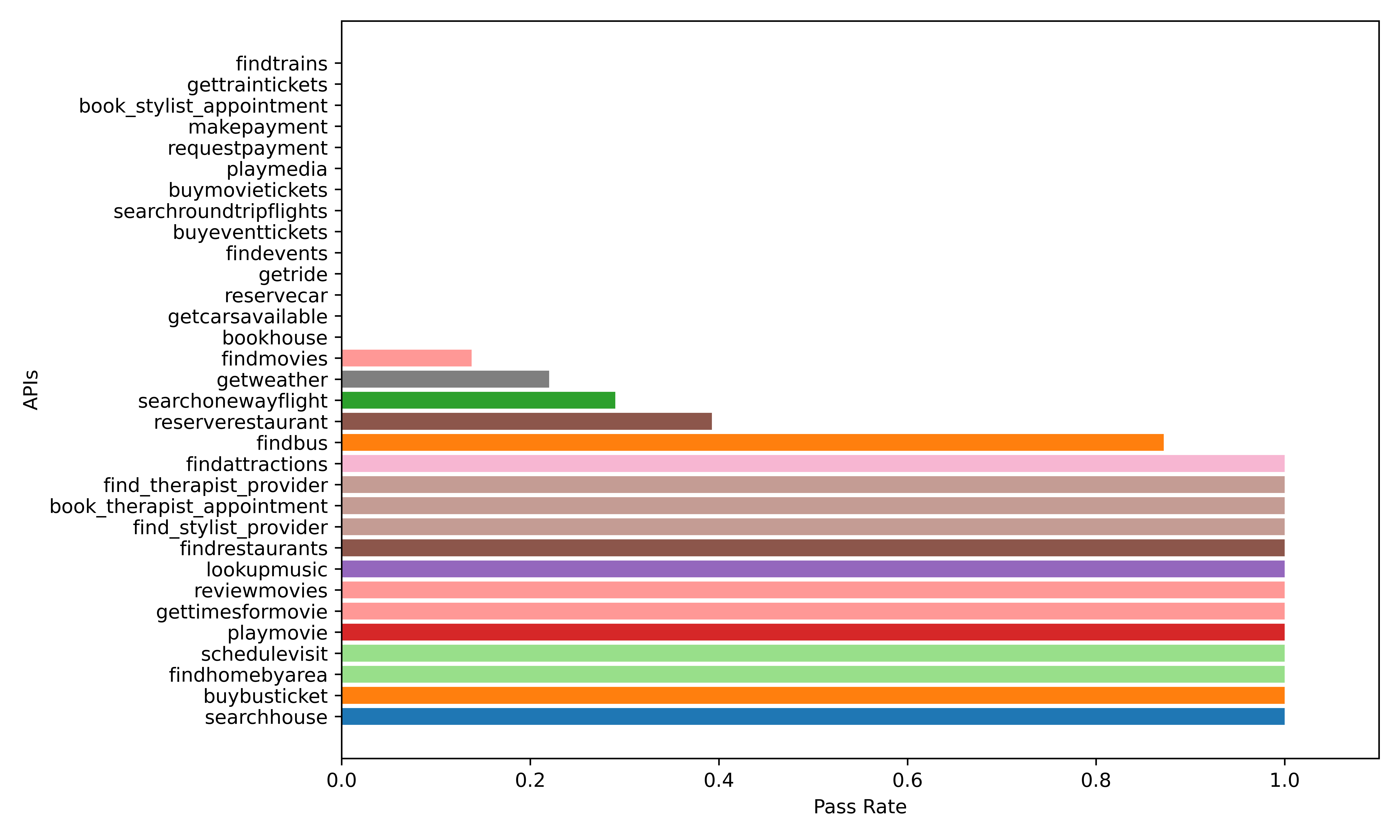}
        \caption{QWen1.5-72B}
        \label{fig:tc-qwen1.5-72b}
    \end{subfigure}

    \vskip\baselineskip

    \begin{subfigure}[b]{0.3\textwidth}
        \centering
        \includegraphics[width=\textwidth]{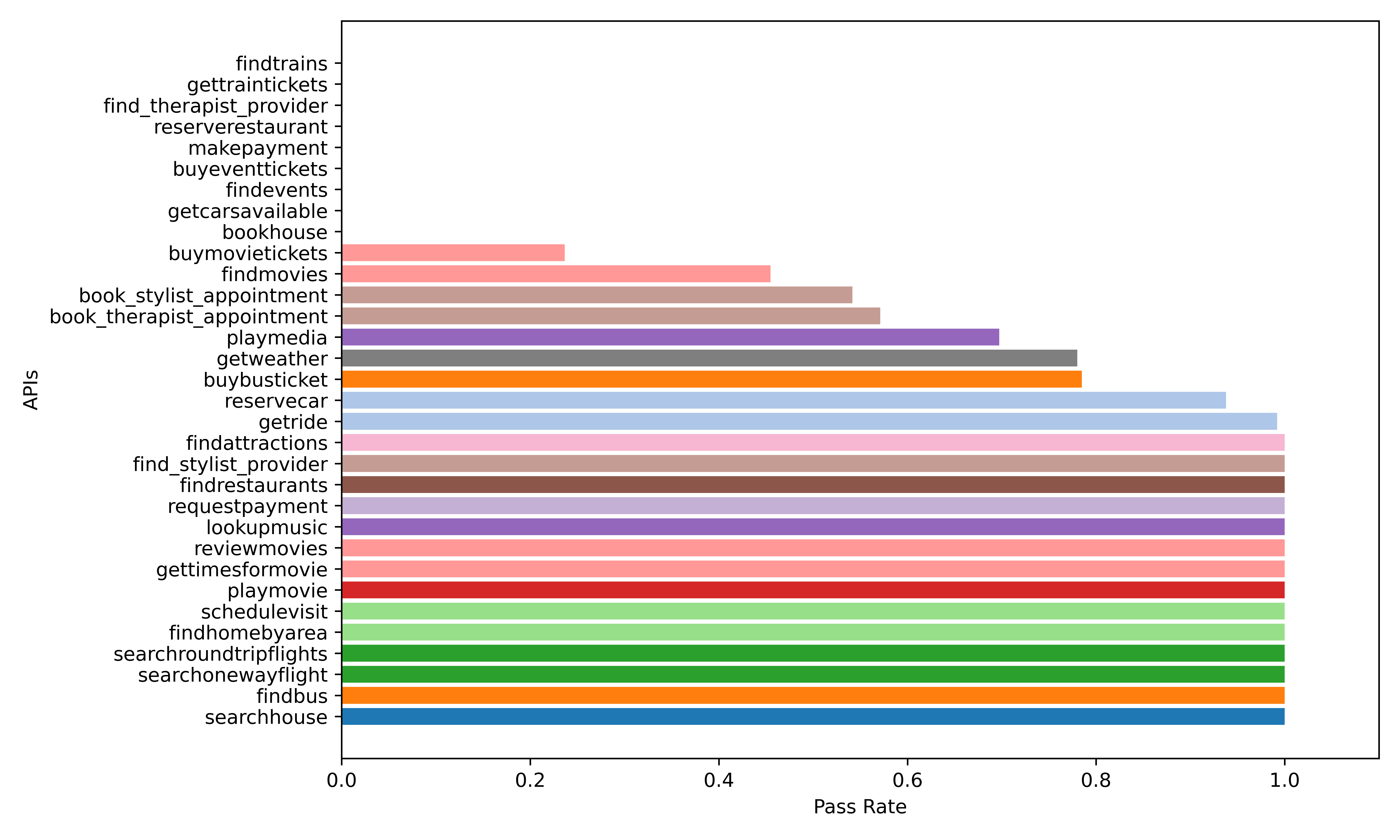}
        \caption{LLaMA3-8B}
        \label{fig:tc-llama3-8b}
    \end{subfigure}
    \hfill
    \begin{subfigure}[b]{0.3\textwidth}
        \centering
        \includegraphics[width=\textwidth]{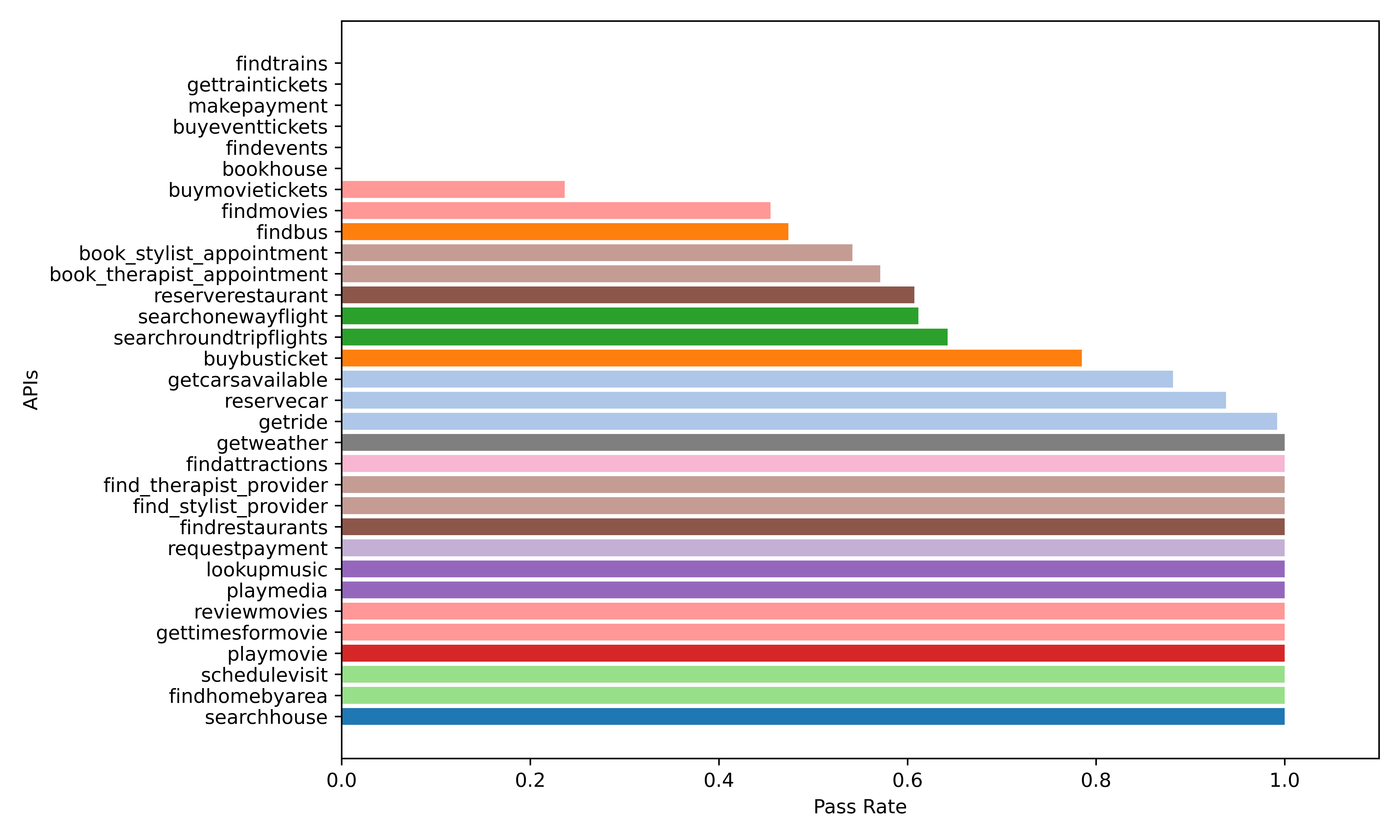}
        \caption{LLaMA3-70B}
        \label{fig:tc-llama-70b}
    \end{subfigure}
    \hfill
    \begin{subfigure}[b]{0.3\textwidth}
        \centering
        \includegraphics[width=\textwidth]{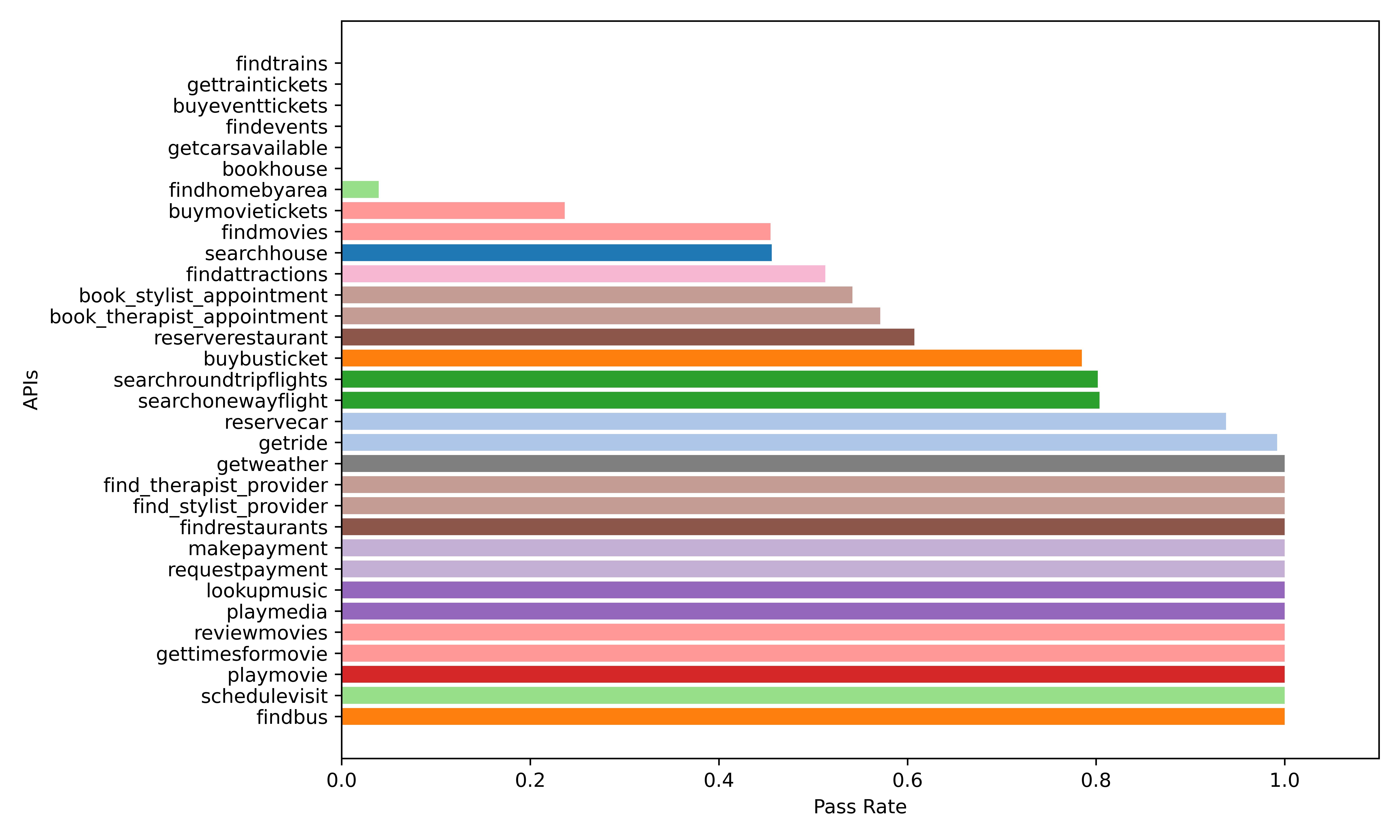}
        \caption{GPT-3.5}
        \label{fig:tc-gpt-3.5}
    \end{subfigure}

    \caption{The tool creation performance of different LLMs on each API. We use same colour to indicate the API comes from same App.}
    \label{fig:appendix_tool_creation}
\end{figure*}

\subsection{Tool Selection}

\begin{figure*}[t]
    \centering
    \begin{subfigure}[b]{0.3\textwidth}
        \centering
        \includegraphics[width=\textwidth]{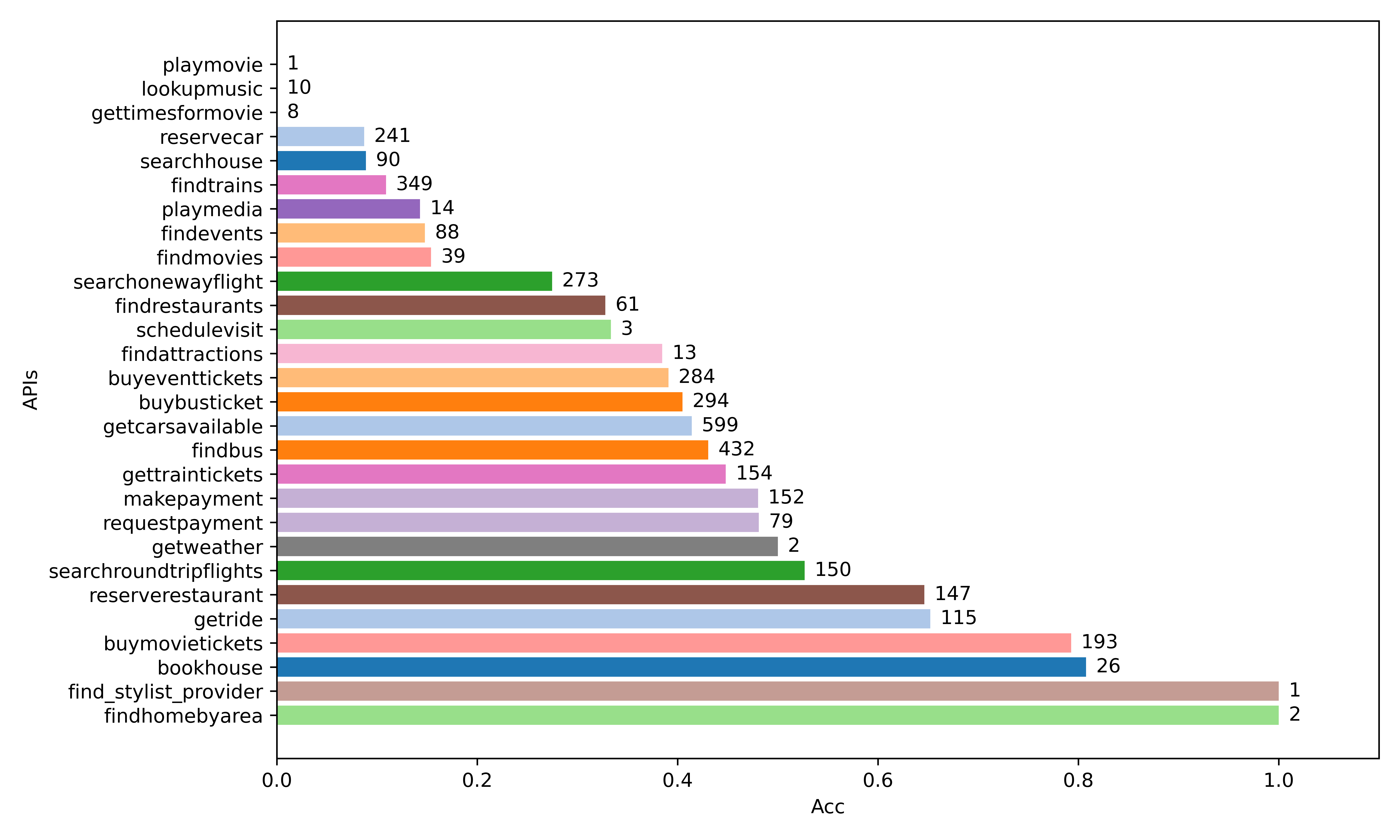}
        \caption{LLaMA2-7B}
        \label{fig:llama2-7b}
    \end{subfigure}
    \hfill
    \begin{subfigure}[b]{0.3\textwidth}
        \centering
        \includegraphics[width=\textwidth]{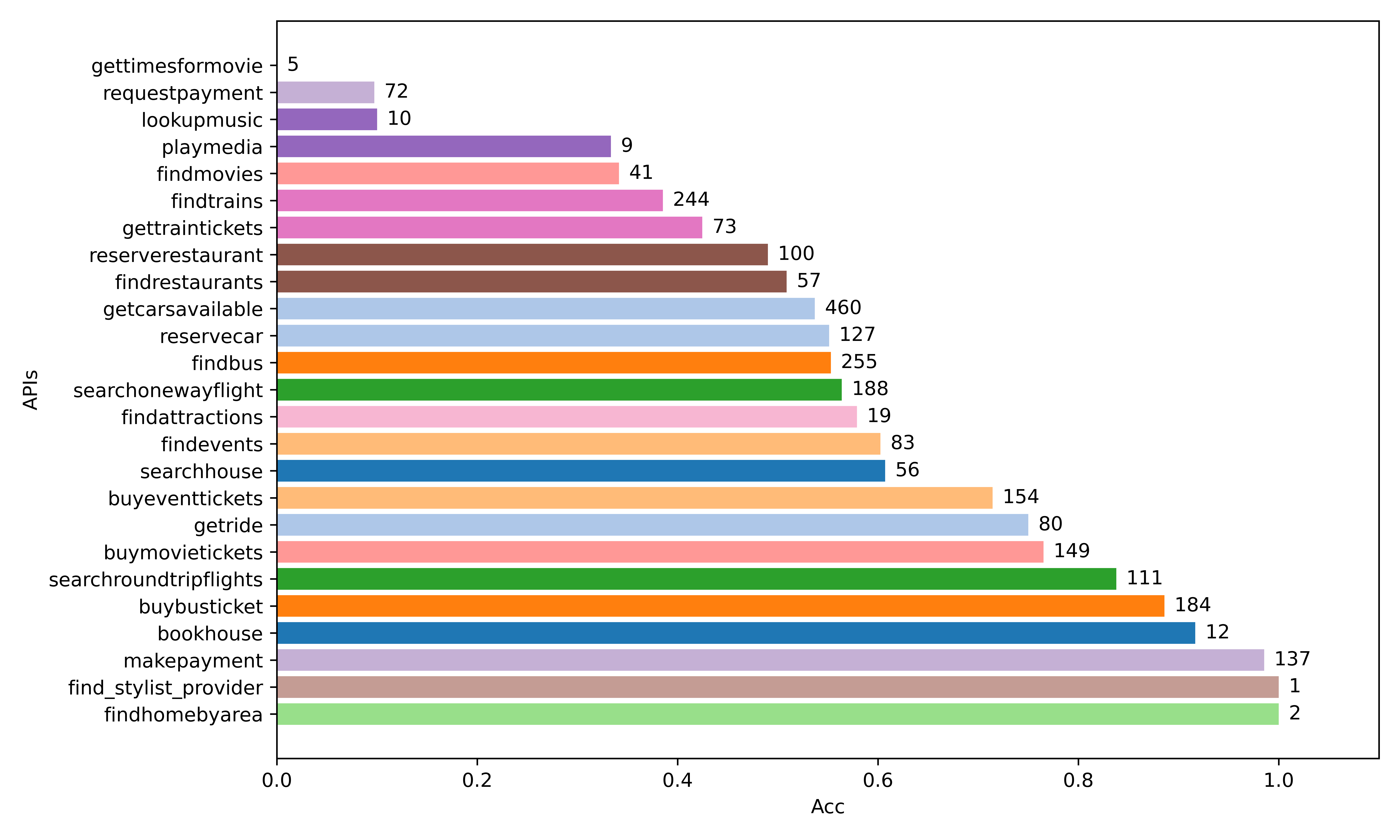}
        \caption{LLaMA2-13B}
        \label{fig:llama2-13b}
    \end{subfigure}
    \hfill
    \begin{subfigure}[b]{0.3\textwidth}
        \centering
        \includegraphics[width=\textwidth]{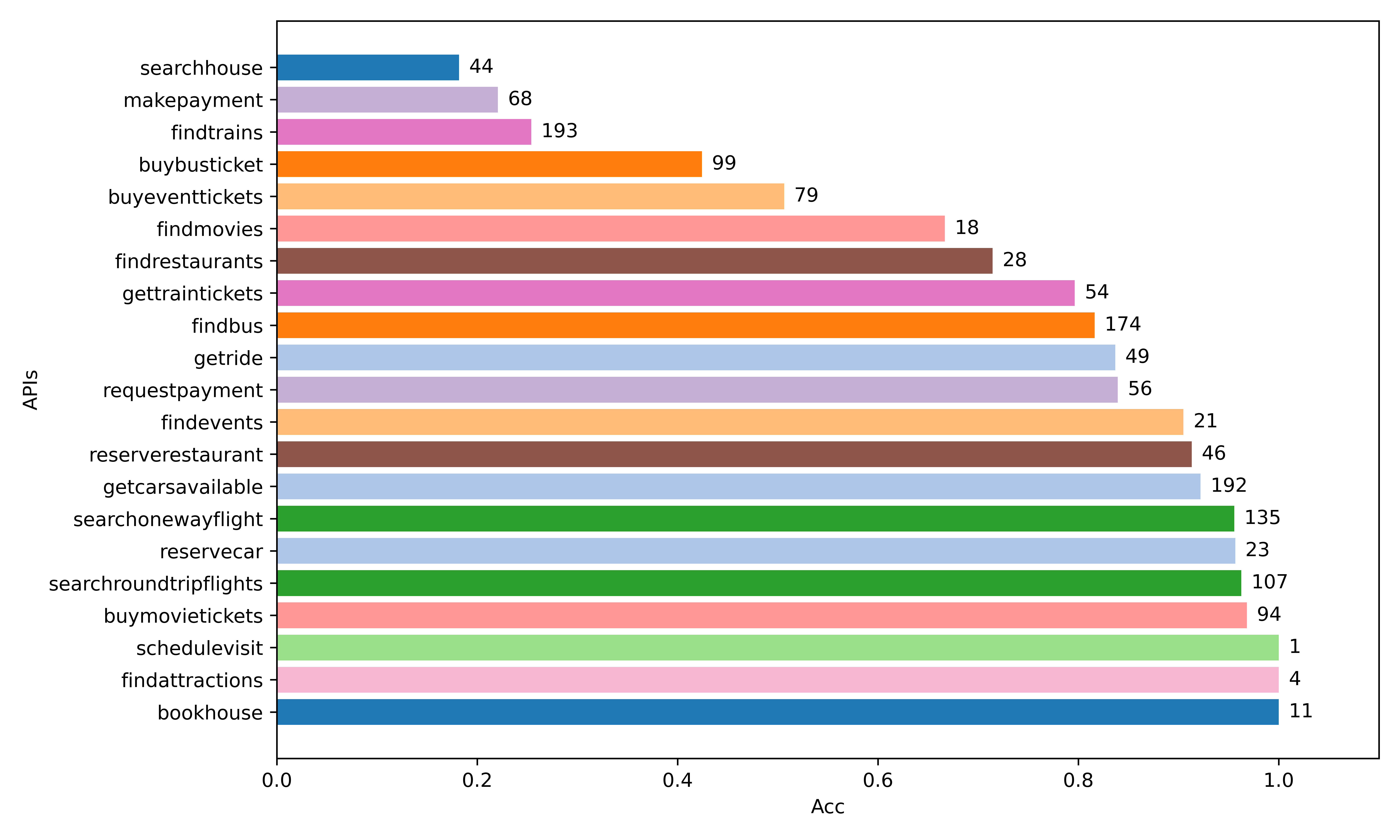}
        \caption{LLaMA2-70B}
        \label{fig:llama2-70b}
    \end{subfigure}

    \vskip\baselineskip

    \begin{subfigure}[b]{0.3\textwidth}
        \centering
        \includegraphics[width=\textwidth]{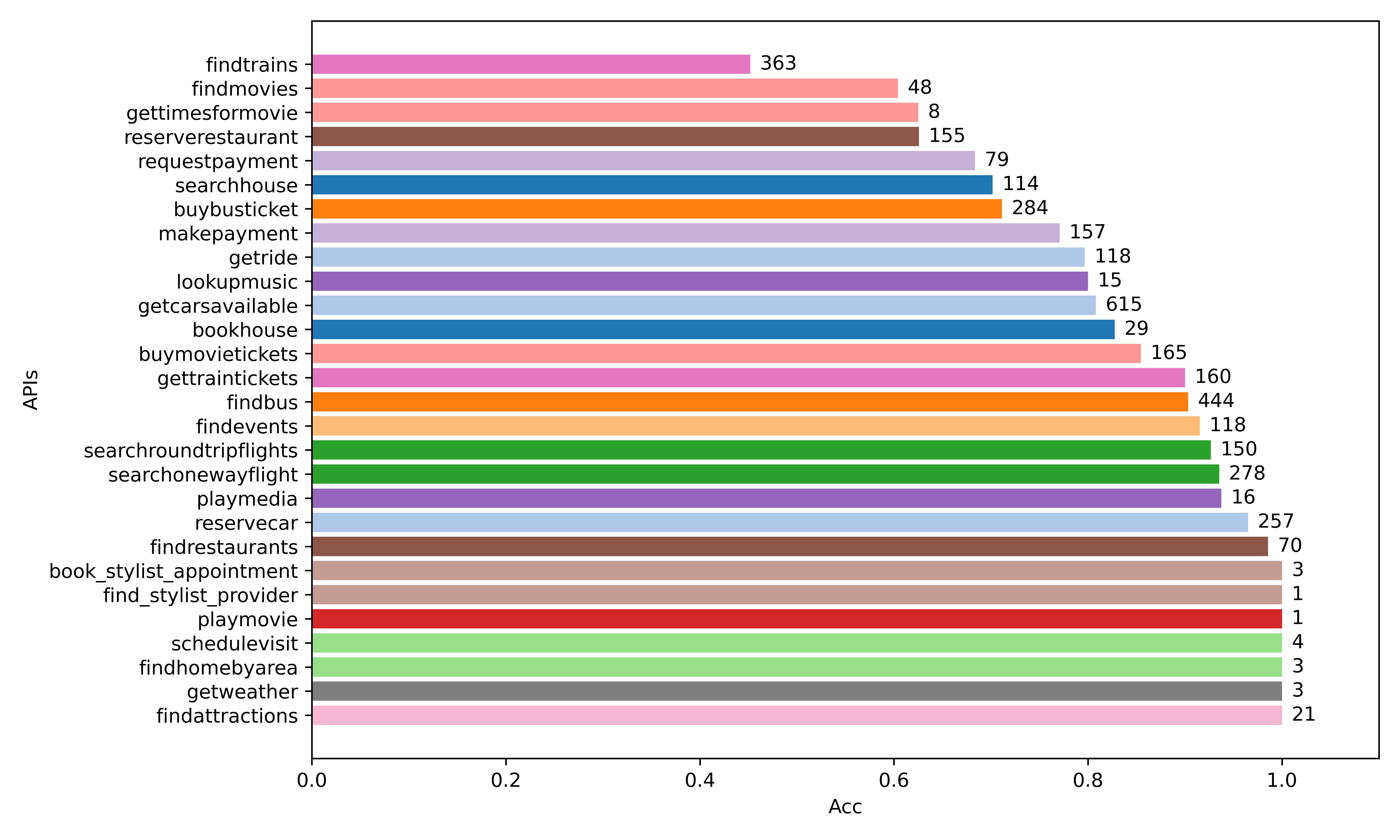}
        \caption{QWen1.5-7B}
        \label{fig:qwen1.5-7b}
    \end{subfigure}
    \hfill
    \begin{subfigure}[b]{0.3\textwidth}
        \centering
        \includegraphics[width=\textwidth]{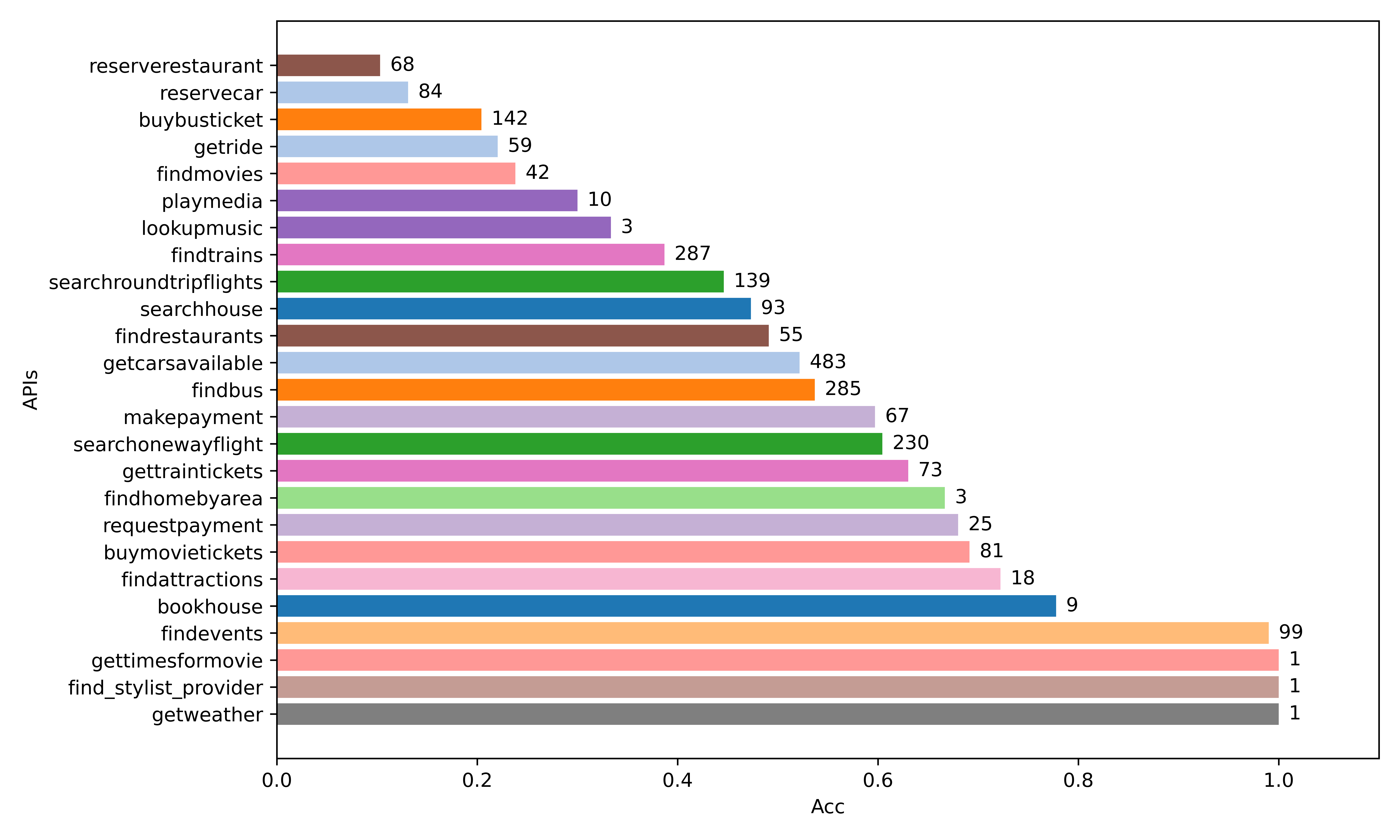}
        \caption{QWen1.5-14B}
        \label{fig:qwen1.5-14b}
    \end{subfigure}
    \hfill
    \begin{subfigure}[b]{0.3\textwidth}
        \centering
        \includegraphics[width=\textwidth]{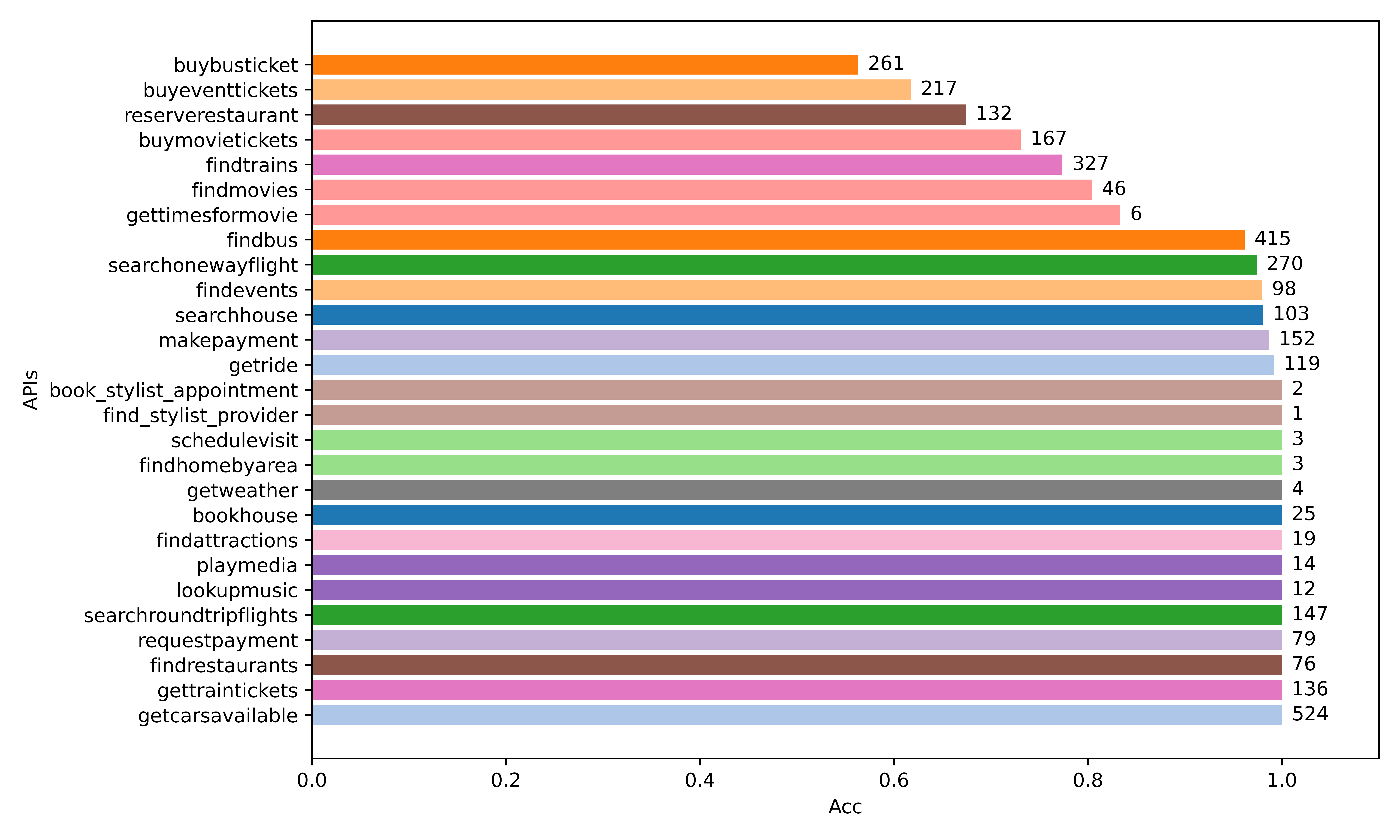}
        \caption{QWen1.5-72B}
        \label{fig:qwen1.5-72b}
    \end{subfigure}

    \vskip\baselineskip

    \begin{subfigure}[b]{0.3\textwidth}
        \centering
        \includegraphics[width=\textwidth]{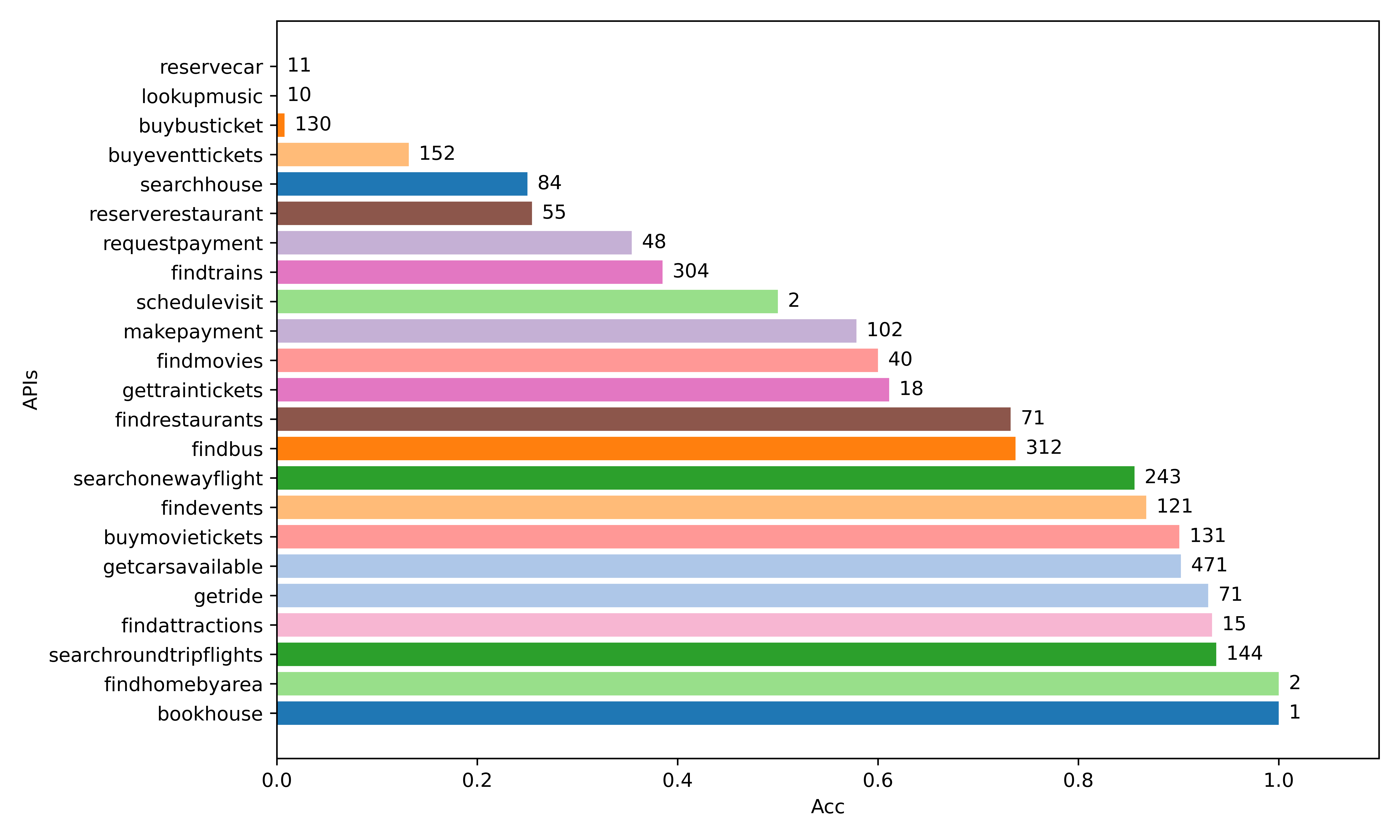}
        \caption{LLaMA3-8B}
        \label{fig:llama3-8b}
    \end{subfigure}
    \hfill
    \begin{subfigure}[b]{0.3\textwidth}
        \centering
        \includegraphics[width=\textwidth]{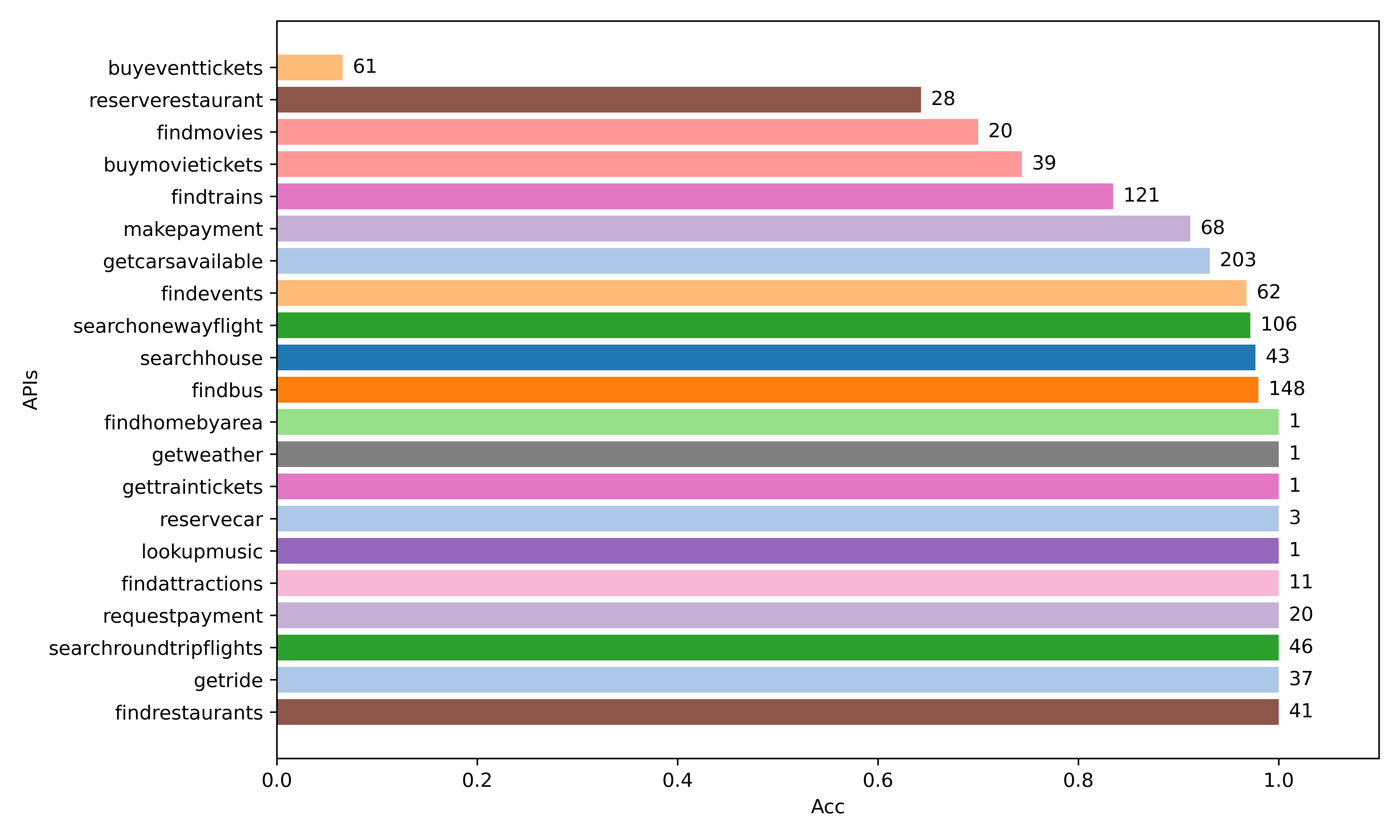}
        \caption{LLaMA3-70B}
        \label{fig:llama-70b}
    \end{subfigure}
    \hfill
    \begin{subfigure}[b]{0.3\textwidth}
        \centering
        \includegraphics[width=\textwidth]{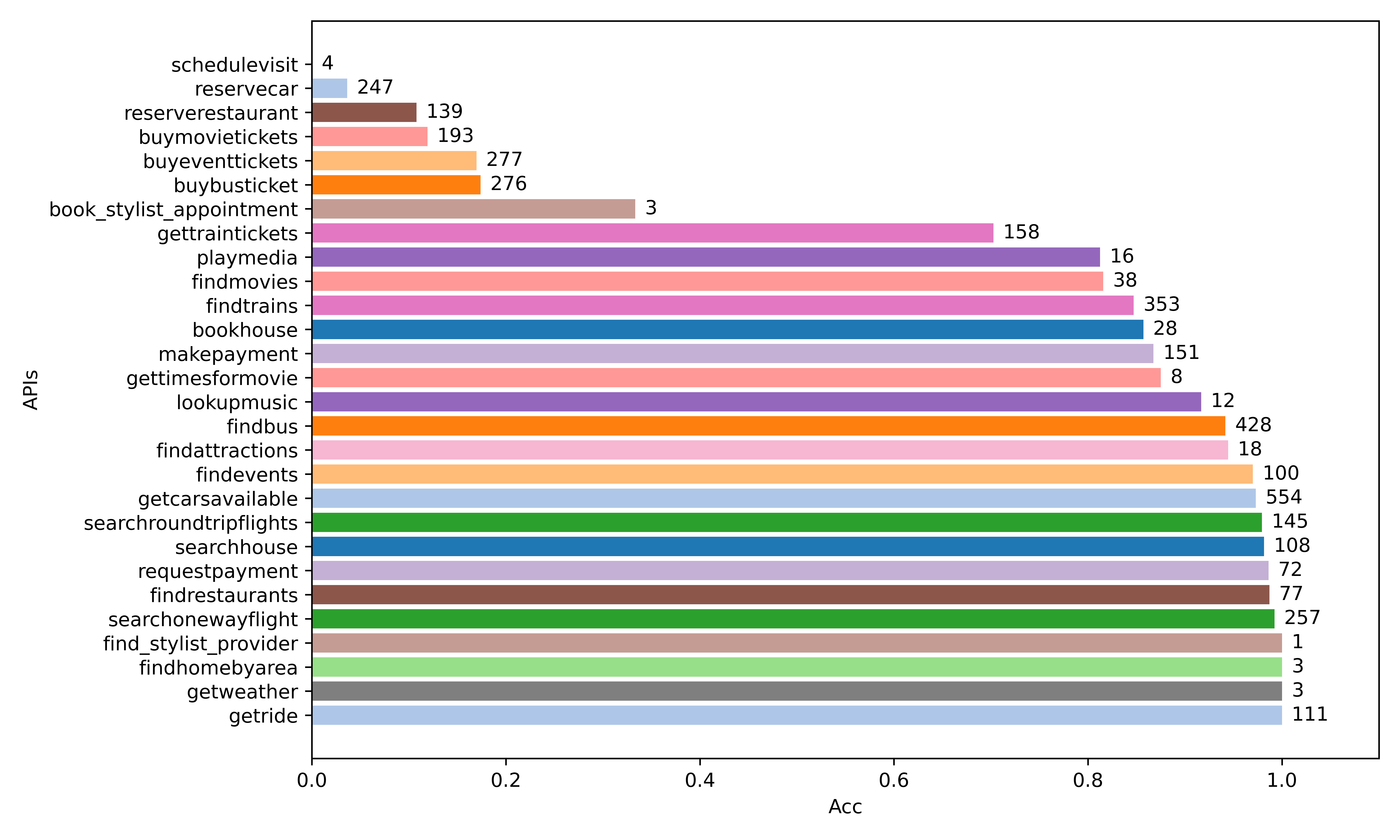}
        \caption{GPT-3.5}
        \label{fig:gpt-3.5}
    \end{subfigure}

    \caption{The performance of different LLMs on each API. We use same colour to indicate the API comes from same App. We also provide the frequency information at the end of bar.}
    \label{fig:appendix_tool_selection}
\end{figure*}

To explore the specfic performance of different LLMs on each API, we provide the accuracy of each API for each LLM as shown in Figure~\ref{fig:appendix_tool_selection}. We could draw several conclusion from the results.

\textbf{\textit{Generally, as the model size increases, it usually leads to better results, as validated in LLaMA2 and LLaMA3 series models.}} However, we find that QWen1.5-7B and QWen1.5-72B surprisingly achieves better performance considering their sizes and QWen1.5-14B is the worst. Specifically, \textbf{\textit{most of LLMs achieves higher performance at \texttt{Homes}, \texttt{Hotels} and \texttt{Travel} Apps, and lower performance at \texttt{Events}, \texttt{Restaurants}, and \texttt{Rents} Apps.}} We attribute this to there are several confusing APIs in later Apps. For example, \texttt{getcaravaiable} and \texttt{reservecar} in the \texttt{Rents} App, the agent usually needs to call \texttt{getcaravaiable} first and then \texttt{reservecar} to fulfill the use task, however, these APIs share most of common arguments and agent may misunderstand the relationship across them. Furthermore, we also analyze the App of unnecessary API calls, and we find most of them comes from \texttt{Buses}, \texttt{Rents} and \texttt{Trains}, the most unnecessary API calls are \texttt{getcarsavailable} and \texttt{buybusticket}.

\subsection{Tool Execution}

\begin{table*}[!t]
\setlength{\belowcaptionskip}{0pt}
    \centering
    \begin{adjustbox}{max width=1.0 \textwidth}
    \begin{tabular}{l | ccc}
    \toprule
    \textbf{Model} & \textbf{Miss} ($\downarrow$) & \textbf{Extra} ($\downarrow$) & \textbf{Value Mismatch} ($\downarrow$)  \\
    \hline
    LLaMa3-8B & 13.2 & 3.9 & 29.3 (65.7, 11.9) \\
    LLaMA2-70B & 19.0 & 27.6 & 41.6 (52.9, 15.4) \\
    LLaMA3-70B & 8.9 & 3.7 & 27.1 (70.6, 8.3)  \\
    \hdashline
    GPT-3.5 & 18.5 & 4.0 & 23.5 (55.8, 20.6) \\
    GPT-4o & 12.4 & 0.9 & 24.3 (58.3, 16.3) \\
    \bottomrule
    \end{tabular}
    \end{adjustbox}
    \caption{Error Analysis of Tool Execution. It is worthy noting the Missing (Extra) column stands for the percentage of missing (extra) keys between predicted arguments and ground truth arguments. We also indicate two major types of value mismatch in (date-related mismatch, location-related mismatch).}
    \label{tab:appendix_execution}
\end{table*}

Besides matching ans missing analysis at the main experiments, we additionally provide specific analysis for keys and values in the arguments when there exists both predicted arguments and ground truth arguments under the same API. In other words, we do not consider the determined API to be wrong or empty and require the keys and values to be exact matches. Table~\ref{tab:appendix_execution} shows the final results.

We firstly emphasize that it is relatively unfair to directly compare performance across different LLMs since the total number of samples is different due to the different setting. However, we can observe several trends: \textbf{\textit{Keys.}} It is obvious that it is challenging for existing LLMs to recognize all arguments from the multi-turn dialogue even given all descriptions about the APIs. Almost all LLMs tend to miss some arguments except LLaMA2-70B, resulting in higher missing errors. Upon further investigation, we've identified a pattern in the missing cases: they tend to occur when there are optional arguments that the agent fails to predict. Additionally, this issue arises when the agent predicts extra arguments, as it may assign default values to certain optional arguments, such as the current date, or assume the number of passengers is always one.

\textbf{\textit{Values.}} Most of value mismatch comes from the time and location related keys, such as \textit{pickup\_time} and \textit{from}. The timing issue primarily stems from incorrect formatting and erroneous reasoning based on the given current date. On the other hand, location issues are often due to commonsense knowledge errors or hallucinations. For example, the dialogue agent might predict an incorrect location or use commonly known aliases for cities (e.g., "NYC" for New York City, "LAX" for Los Angeles). These observations requires more attention to further improve the performance of dialogue agent.

\subsection{Human Evaluation}
\label{human_study}
We hire three well-educated master students and randomly sample 50
response for each model. They were then asked to assign a score to each response, ranging from 1 (extremely poor, such as totally unrelated or API error) to 5 (extremely good, such as all details about the arguments are clearly stated). We provided specific examples for each score to illustrate the subtle differences in assignment. We calculate the average score of these annotators.

\end{document}